\documentclass[sigconf, nonacm]{acmart}

\usepackage{float}
\usepackage{placeins}
\usepackage{tcolorbox} 
\usepackage{listings}
\usepackage{subcaption}
\usepackage{tabularx} 
\usepackage{graphicx} 
\usepackage{amssymb}  
\usepackage{multirow}
\usepackage{mdwlist}
\usepackage{balance}
\usepackage{booktabs}
\usepackage{adjustbox}
\usepackage{threeparttable}
\usepackage{makecell}   
\usepackage{bbding}

\usepackage{algorithm}
\usepackage{algpseudocode}
\usepackage{amsmath}

\usepackage{tcolorbox}
\tcbuselibrary{skins, breakable} 
\usepackage{xcolor}
\usepackage{caption}

\newcommand{\framesectiontitle}[1]{\textbf{\textcolor{myPurple}{\underline{#1}}}}

\definecolor{myRed}{rgb}{0.8, 0.0, 0.0}
\definecolor{myPurple}{rgb}{0.4, 0.0, 0.6}
\definecolor{myBlue}{rgb}{0.0, 0.2, 0.8}
\definecolor{myGreen}{rgb}{0.0, 0.6, 0.3}
\definecolor{myOrange}{rgb}{1.0, 0.4, 0.0}
\definecolor{myGray}{gray}{0.5}
\definecolor{myLightGray}{gray}{0.85}
\definecolor{myTeal}{rgb}{0.0, 0.5, 0.5}
\definecolor{myPink}{rgb}{0.9, 0.1, 0.4}
\definecolor{lightYellow}{RGB}{255, 255, 204} 

\newcommand{\redhighlight}[1]{\textcolor{myRed}{\textbf{#1}}}

\tcbset{
    mybluebox/.style={
        colback=blue!5!white, 
        colframe=blue!75!black, 
        boxrule=0.5mm, 
        arc=4mm, 
        width=\textwidth, 
        sharp corners,
        fonttitle=\bfseries,
    }
}

\newcommand\vldbpagestyle{plain} 

\begin{document}
\title{SING-SQL: A Synthetic Data Generation Framework for In-Domain Text-to-SQL Translation}

\author{Hasan Alp Caferoğlu}
\affiliation{%
  \institution{Bilkent University}
  \streetaddress{}
  \city{Ankara}
  \state{Turkey}
  \postcode{06800}
}
\email{alp.caferoglu@bilkent.edu.tr}

\author{Mehmet Serhat Çelik}
\affiliation{%
  \institution{Bilkent University}
  \streetaddress{}
  \city{Ankara}
  \state{Turkey}
  \postcode{06800}
}
\email{serhat.celik@ug.bilkent.edu.tr}

\author{Özgür Ulusoy}
\affiliation{%
  \institution{Bilkent University}
  \streetaddress{}
  \city{Ankara}
  \country{Turkey}
  \postcode{06800}
}
\email{oulusoy@cs.bilkent.edu.tr}


\begin{abstract}
Translating natural language questions into SQL has become a core challenge in enabling non-technical users to query databases. While recent work has explored large-scale synthetic data generation to improve model performance through post-training, most efforts emphasize cross-domain generalization. This leaves a gap for real-world enterprise scenarios, where models need to specialize to a single database schema and organizations require to be able to evaluate their Text-to-SQL systems on their own databases. To address this, we introduce \textbf{SING-SQL}\textsuperscript{1}, a fully automated two-stage framework for generating high-quality, high-coverage synthetic Text-to-SQL data for any target database, without relying on SQL logs or manual annotations. Our approach hierarchically partitions a database schema into sub-schemas, synthesizes SQL queries across multiple complexity levels, and applies a quality-aware pipeline that includes LLM-as-a-judge validation, executability checks, automatic repair, and column balancing. We further release \textbf{SingSQL-LM}, a family of compact language models fine-tuned on the synthetic data, achieving strong in-domain generalization. On the subset of the BIRD benchmark, SingSQL-LM-3B-R64 reaches 82.87\% Soft F1 and 73.03\% EX upper bound with 32 candidates, outperforming the best 3B-scale baseline by +16.21 in Soft F1 and +12.36 in EX. At the 1.5B scale, SingSQL-LM-1.5B-R64 improves over prior systems by +9.30 in Soft F1 and +4.49 in EX. On synthetic evaluation sets, SingSQL-LMs exceed prior systems by wide margins, establishing state-of-the-art performance among open models at comparable scales. Moreover, our study of context management strategies reveals that schema-free fine-tuning combined with schema-only inference provides the most robust results. Together, these findings establish SING-SQL as a scalable, database-agnostic paradigm for producing and evaluating enterprise-grade Text-to-SQL systems.
\end{abstract}

\maketitle

\pagestyle{\vldbpagestyle}


\footnotetext[1]{Artifacts have been made available at \url{https://github.com/HasanAlpCaferoglu/SING-SQL}.}

\section{Introduction}
\label{sec:introduction}

Translating natural language queries into structured query language (Text-to-SQL) has become a key area of interest for both academia and industry, as it enables non-technical users to query relational databases using natural language. Typical Text-to-SQL systems consist of several core components, including schema linking, SQL generation, SQL correction, and SQL selection. To improve performance at each stage of this workflow, fine-tuning large language models (LLMs) on task-specific data has become a common practice. However, such improvements often come at the cost of obtaining large-scale, high-quality annotated datasets, which are expensive and time-consuming to collect.

While prior work has proposed frameworks to synthetically generate labeled Text-to-SQL data ~\cite{li-2024-codes, li2025omnisqlsynthesizinghighqualitytexttosql,guo-etal-2025-sqlforge, yang-etal-2024-synthesizing-t2s-data-from-weak-and-strong-llms, cheng_etal_sqlord}, most focus on the cross-domain setting—targeting generalization across diverse and unrelated databases. However, this setting does not reflect the needs of many real-world applications, where models are deployed in narrowly scoped environments and are expected to serve a specific organization operating on one or a few domain-specific databases. For example, an e-commerce company is unlikely to require querying a Formula 1 database, but instead needs highly accurate query generation over its internal, proprietary schema. In such scenarios, domain-specialized models trained with high-coverage in-domain data are far more beneficial than broadly generalized models.

To address this need, we introduce \textbf{SING-SQL} (Synthetic IN-domain Data Generation for Text-to-SQL), a framework designed to generate high-quality synthetic Text-to-SQL data tailored to any specific relational database. In addition, we present \textbf{SingSQL-LM}, a family of compact language models fine-tuned on the data produced by SING-SQL. We evaluate our approach on a subset of the BIRD benchmark as well as on our own synthetically generated in-domain dataset. Experimental results demonstrate that SING-SQL enables the creation of large volumes of high-quality, high-coverage domain-specific Text-to-SQL pairs. Such data can be leveraged by enterprises to fine-tune internal language models or to evaluate the performance of agentic systems operating in Text-to-SQL workflows.

In SING-SQL, we design a two-stage synthetic data generation framework. In the first stage, the target database is systematically partitioned into a diverse collection of sub-schemas, each defined by a unique combination of tables and column subsets. This decomposition is performed at two levels: first, at the table level, where joinable subsets of tables are selected; and second, at the column level, where a sliding window strategy is applied over non-key columns to generate multiple column-wise variations for each table. This hierarchical partitioning enables fine-grained control over the schema context of each example, reduces generation noise, and ensures semantic alignment between input questions and the underlying schema. By systematically varying both table and column configurations, the process promotes full database coverage across the synthesized dataset while keeping the generation process tractable and interpretable. In the second stage, we generate SQL queries for each sub-schema across multiple complexity levels—simple, moderate, challenging, and window—and translate them into natural language questions. To ensure data quality, we apply a series of filtering steps: first, an LLM-as-a-judge module evaluates the logical validity and semantic alignment of each SQL–Text pair; next, we test the executability of each SQL query and invoke automatic repair for non-executable cases when possible. For the subset of validated and executable examples, we instruct the model to generate step-by-step reasoning traces using a divide-and-conquer prompting strategy, enhancing interpretability and instructional value. Finally, to address schema imbalance, we perform a second round of column-focused SQL generation, targeting underrepresented columns based on usage frequency thresholds. Together, these components produce a high-quality, high-coverage synthetic dataset tailored for in-domain Text-to-SQL tasks. 

Our key contributions can be summarized as follows:

\begin{itemize}
\item  We propose \textbf{SING-SQL}, a two-stage synthetic data generation framework designed for in-domain Text-to-SQL tasks. It generates high-quality, high-coverage SQL–Text pairs tailored to any specific relational database.

\item We introduce a \textbf{hierarchical sub-schema construction strategy} that partitions the database at both table and column levels, enabling controllable and semantically meaningful data generation while ensuring comprehensive schema coverage.

\item With \textbf{SING-SQL}, we present a \textbf{quality-aware SQL–Text generation pipeline} that incorporates complexity-controlled SQL synthesis, LLM-as-a-judge validation, executability checks, automatic SQL repair, and reasoning trace generation.

\item We address schema imbalance through a \textbf{column-focused generation step}, which targets underrepresented columns using frequency-based thresholds to improve column-level data coverage.

\item We release \textbf{SingSQL-LM}, a family of compact language models fine-tuned on the synthetic data produced by SING-SQL, and demonstrate their effectiveness on a subset of the BIRD benchmark and in-domain synthetic datasets.
\end{itemize}

\section{Related Work}

\label{sec:related_work}
\subsection{Synthetic Data Generation for Text-to-SQL}
Large language models (LLMs) for Text-to-SQL often benefit from fine-tuning on curated datasets, either drawn from existing benchmarks~\cite{Pourreza-2024-dts-sql, pourreza2025chasesql} or synthesized to expand coverage across domains. A series of recent works propose diverse strategies for generating synthetic training data.

CODES~\cite{li-2024-codes} constructs data from two sources: SQL-related corpora and NL-to-code datasets. It introduces a bi-directional augmentation procedure. In the question-to-SQL direction, proprietary LLMs generate user-like questions from real examples, followed by SQL generation. In the SQL-to-question direction, SQL queries are paired with template-based questions, which are then rephrased by LLMs to improve naturalness. Yang et al. ~\cite{yang-etal-2024-synthesizing-t2s-data-from-weak-and-strong-llms} synthesizes training data in two complementary phases. First, strong proprietary LLMs are used to synthesize diverse, high-quality Text-to-SQL pairs across domains, thereby avoiding over-representation of specific schemas. This strong data is used to enhance the base model's Text-to-SQL translation capabilities. Next, through preference learning, weak data is generated from smaller LLMs by sampling candidate SQLs, labeling them through execution agreement with gold queries, and applying Direct Preference Optimization (DPO) on positive/negative pairs. OMNISQL\cite{li2025omnisqlsynthesizinghighqualitytexttosql} scales up data generation by first synthesizing realistic databases from web tables via LLMs, ensuring broad domain coverage. SQL queries are then produced with complexity-aware strategies to balance query difficulty, followed by back-translation into natural questions. To enrich linguistic diversity, questions are paraphrased in different language styles. Additionally, chain-of-thought (CoT) rationales\cite{kojima-2022-LLMS-are-zero-shot-reasoners, wei2022COT} are generated alongside each question-SQL pair to provide auxiliary training signals and interpretability. A synthetic Text-to-SQL framework SQLForge ~\cite{guo-etal-2025-sqlforge} focus on the reliability and diversity of the generated data in four-stage. Initially, An AST-based SQL Parser turns seed queries into grammar-preserving templates and enriches them by subtree crossover. Secondly, SQL foundry stage, iteratively explores new domains by jointly proposing domain names with auxiliary SQL to prevent domain drift. Then, for the synthesized SQL statements, corresponding database schema expression is generated via schema architect stage. Finally, a schema-aware reverse translator produces human-readable questions that reflect the SQL’s intent. SQLord~\cite{cheng_etal_sqlord} addresses data scarcity by exploiting developer-authored SQL queries and their comments. A model called RevLLM is trained on a small set of SQL–comment pairs to generate natural questions for SQL queries. Using RevLLM, large-scale pseudo-labeled question–SQL pairs are synthesized, which in turn train a specialized Text-to-SQL model. Shkapenyuk et al.~\cite{shkapenyuk2025automaticmetadataextractiontexttosql} take a different angle by focusing on metadata extraction through profiling, query log analysis, and SQL-to-text generation. 

Most prior works primarily focus on cross-domain data synthesis to improve generalization across databases. However, in enterprise contexts, organizations often require models specialized for their own databases. Recent efforts move in this direction. TailorSQL~\cite{vaidya2025tailorsqlnl2sqltailoredquery} constructs a workload-aware retrieval corpus by synthesizing documents from both schema and query logs, using LLMs to generate natural-language questions conditioned on SQL and schema. However, it primarily improves performance through prompt augmentation rather than large-scale data generation and post-training. Having a similar objective to our work, SelectCraft~\cite{chafix_selectcraft} introduces a domain-specific data generation framework that synthesizes SQL queries by mimicking real-world query distributions over existing databases. Its controllability is limited to distributional aspects of SQL components (e.g., operators, conditions, join counts and types), but it does not enable fine-grained control over database elements such as tables or columns, nor does it address scalability to large and complex schemas. In parallel, SiriusBI~\cite{SiriusBI} introduces an automated pipeline for domain-specific data generation, with optional manual verification to further enhance quality assurance. The system aims to mitigate the sharp performance drops often observed when models are transferred across domains with limited generalization capacity. However, SiriusBI primarily leverages SQL query logs, which are not always available, and it lacks mechanisms for fine-grained control over schema elements. As a result, while it provides a practical solution for reducing domain transfer bottlenecks, its capacity for scaling to complex schemas or enforcing systematic schema coverage remains limited.

Our work, SING-SQL, addresses this gap by enabling enterprises to curate Text-to-SQL data tailored to their schema without relying on prior workloads. It enforces comprehensive schema coverage through hierarchical sub-schema construction and targeted column balancing, while ensuring complexity-controlled data via quality-aware SQL–Text synthesis, thereby moving beyond the cross-domain paradigm.

\subsection{LLM Post-Training for Text-to-SQL}
Out-of-the-box, open-source LLMs generally underperform proprietary counterparts on Text-to-SQL benchmarks. Recent work shows, however, that with targeted post-training strategies using synthetic data, open-source models can achieve competitive or even superior performance.

Supervised fine-tuning (SFT) has been the most widely used method for adapting models to Text-to-SQL corpora~\cite{Pourreza-2024-dts-sql, Gao-2024-dail-sql, wang-2024-macsql, talaei2024chess, pourreza2025chasesql, li2025omnisqlsynthesizinghighqualitytexttosql, XiYanSQL, sheng_2025_basesql}. Beyond SFT,  direct preference optimization (DPO), a preference learning technique, has been utilized to further align models with desired behaviors~\cite{rafailov_2023_dpo, yang-etal-2024-synthesizing-t2s-data-from-weak-and-strong-llms}. More recently, reinforcement learning (RL) methods have been explored to strengthen LLM reasoning. Group Relative Policy Optimization (GRPO)\cite{DeepSeekMath} has been applied to incorporate reward signals, demonstrating promising improvements in Text-to-SQL systems~\cite{yao_2025_arctic_text2sql_r1_simplerewardsstrong, pourreza2025reasoningsqlreinforcementlearningsql}. Hybrid post-training pipelines that combine SFT with RL-based optimization also explored to enhance model performance by leveraging both supervised and reward-driven signals~\cite{sheng2025cscsqlcorrectiveselfconsistencytexttosql, sheng2025slmsqlexplorationsmalllanguage, ma_2025_sql_r1, papicchio_2025_think2sql_reinforcellmreasoning}.

Despite these advances, reinforcement learning methods can be computationally expensive and difficult to scale for small enterprise settings. In our work, we adopt a parameter-efficient alternative: supervised fine-tuning with Low-Rank Adaptation (LoRA)~\cite{hu_2022_lora_lowrankadaptation}, enabling scalable specialization of open-source models to enterprise databases without the overhead of RL training.

\subsection{Text-to-SQL Systems}

\subsubsection{Schema Linking}
As the context length and capabilities of LLMs increase, more information can be inserted into prompt. The contextual information becomes important and effect the system performance~\cite{is_long_context_all_you_need}. One of the component of the context for the Text-to-SQL systems is database schema. Although there are works ~\cite{caferoğlu2025esqldirectschemalinking, maamari_2024_schemadeath_distillery} stating the schema filtering is not needed as the the model capabilities and context length increase since models can identify relevant schema component during SQL generation, best performing Text-to-SQL systems ~\cite{pourreza2025chasesql, shkapenyuk2025automaticmetadataextractiontexttosql, XiYanSQL, talaei2024chess, pourreza2025reasoningsqlreinforcementlearningsql} shows the effectiveness of schema filtering and utilize it to achieve high performance. TA-SQL~\cite{qu-etal-2024-ta-sql} leverages initially generated dummy SQL queries to improve schema linking performance. Similarly, with the help of an LLM, Gen-SQL ~\cite{shi-etal-2025-gensql} construct a pseudo-SQL and then pseudo-schema solely based on the question without providing the database schema. Leveraging embedding based retriever to eliminate the irrelevant schema components, pseudo-schema is grounded in actual database schema.  ExSL~\cite{glass2025extractiveschemalinkingtexttosql} proposes a decoder-only approach that reformulates schema linking as an extraction task rather than a generative one. ExSL treats all schema columns as candidates and predicts their relevance using hidden states from the LLM, enabling efficient probability estimation. PSM-SQL ~\cite{yang_2025_psmsqlprogressiveschemalearning} couples multi-granularity semantics with a chain-loop pruning strategy for schema linking. Solid-SQL ~\cite{liu-etal-2025-solidsql} enhances schema linking by parsing gold SQL to obtain supervision and fine-tuning a schema selector with paraphrase-augmented data to improve robustness against synonyms and rephrasings. Schema linking model predictions are integrated into prompts via a “focus on” mechanism that highlights selected tables and columns while retaining full-schema context for fault tolerance. RSL-SQL~\cite{cao_2024_rsl_sql} introduces bidirectional schema linking that unions an LLM-extracted schema with a schema parsed from a preliminary SQL, producing a pruned schema that preserves high recall. For schema linking, CHESS~\cite{talaei2024chess} employs an LLM-based schema selector that filters relevant tables and columns, guided by candidate values retrieved through an LSH-based entity matching mechanism and enriched with column descriptions extracted from a vector database. In our work, we adopt the LSH-based entity matching mechanism from CHESS ~\cite{talaei2024chess} and incorporate synthetic Text-to-SQL data to enhance LLM-based column filtering.

\subsubsection{SQL Generation and Selection}
For a single user question, generating multiple SQL queries is best practice for the Text-to-SQL translation systems in order to improve system performance. ~\cite{wang-2024-macsql, luo2024SuperSQL, lee-2024-mcs-sql, talaei2024chess, pourreza2025chasesql, XiYanSQL, sheng2025cscsqlcorrectiveselfconsistencytexttosql, sheng2025slmsqlexplorationsmalllanguage}. As the number of candidate SQL query increases, the system performance increases upto a point. ~\cite{talaei2024chess,pourreza2025chasesql}. In addition to multiple SQL generation, Text-to-SQL systems contain self-correction ~\cite{caferoğlu2025esqldirectschemalinking, pourreza2025chasesql, pourreza-2023-dinsql, luo2024SuperSQL} module to fix or refine the generated SQL queries to improve the system performance. Self-consistency~\cite{wang2023selfconsistency} is another technique utilized largely by Text-to-SQL systems~\cite{dong-2023-c3, lee-2024-mcs-sql, xie_et_al_opensearch_sql}. In our work, we do not employ self-consistency or self-correction to solely measure the model Text-to-SQL translation capability.

\begin{figure*}[htbp]
  \centering
  \includegraphics[width=1.0\linewidth]{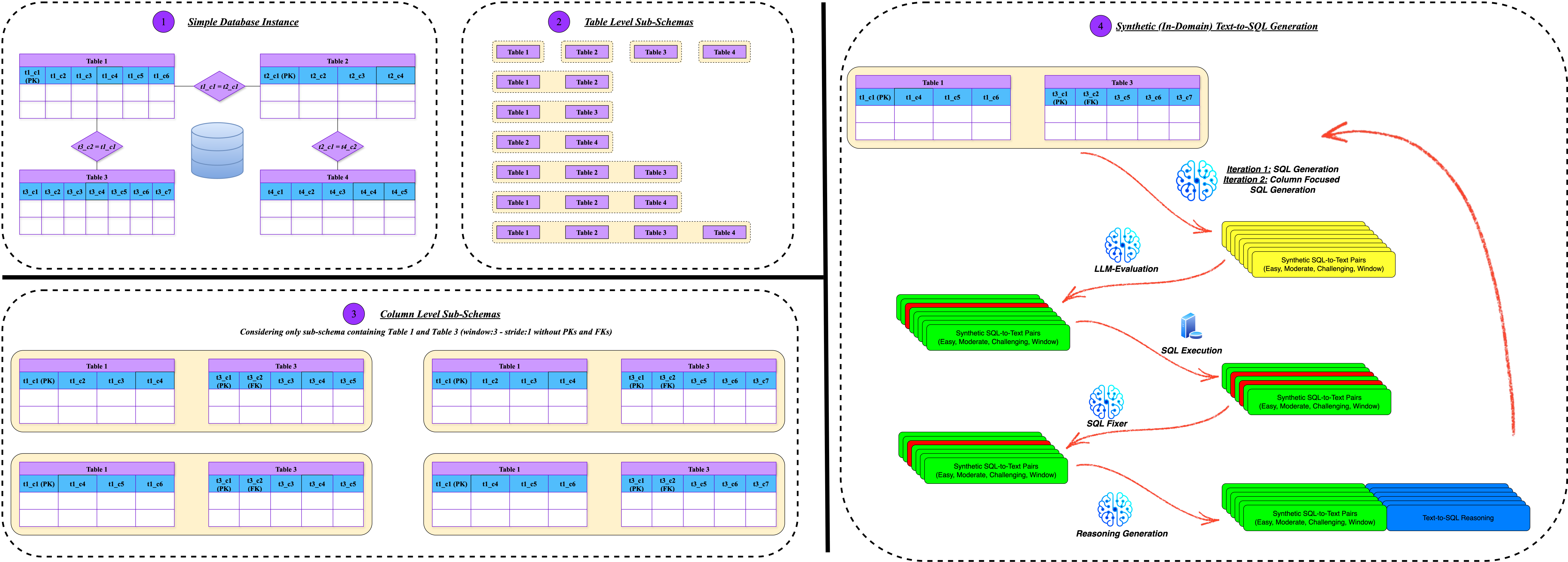}
  \caption{Overview of the Synthetic Data Generation Framework}
  \label{fig:synthetic_t2s_generation}
\end{figure*}

\section{In-Domain Text-to-SQL Synthesis Framework}
\label{sec:methodology_data_synthesis}
To enable comprehensive supervision over all components of a target database, we introduce a two-stage framework for in-domain synthetic Text-to-SQL generation. The first stage involves partitioning the database into a diverse set of sub-schemas, varying in size and structure, to capture different combinations of schema elements. This step ensures that each table and column is systematically included in at least one context. In the second stage, we generate synthetic Text-to-SQL examples conditioned on each sub-schema, enabling fine-grained alignment between natural language questions and the underlying database structure. The resulting dataset achieves high coverage and semantic diversity, enabling more reliable post-training and comprehensive evaluation. The SING-SQL framework is further detailed in Algorithm~\ref{alg:data_gen_algo_overview}.

\begin{algorithm}[htbp]
\caption{General Synthetic Data Generation Algorithm for Single Database}\label{alg:data_gen_algo_overview}
\begin{algorithmic}[1]
\Require window $w$, stride $s$, table counts for sub-schemas $tc$ 
\Ensure \texttt{t2sExamples} 

\State \texttt{subSchemas} $\gets$ \Call{ConstructSubSchemas}{$w$, $s$, $tc$} 
\State \texttt{t2sExamples} $\gets$ \Call{GenT2S}{\texttt{SubSchemas}} 
\State \texttt{columnCounts} $\gets$ \Call{CountCols}{\texttt{t2sExamples}} 
\State\texttt{focusColumns} $\gets$ \Call{GetFocusCols}{\texttt{columnCounts}} 
\Comment{Identify underrepresented or critical columns}
\State \texttt{focusSubSchemas} $\gets$ \Call{FindFocusSchemas}{\texttt{focusColumns}} 
\Comment{Sub-schemas that include the targeted columns}
\State \texttt{colFocusedT2SExamples} $\gets$ \Call{GenT2S}{\texttt{focusSubSchemas}} 
\Comment{Generate more examples focusing on underused columns}
\State \texttt{t2sExamples} $\gets$ \texttt{t2sExamples} $+$ \texttt{colFocusedT2SExamples} 
\State \texttt{t2sExamples} $\gets$ \Call{filter}{\texttt{t2sExamples}} 
\Comment{Removing logically/syntactically incorrect pairs}
\end{algorithmic}
\end{algorithm}

\subsection{Sub-Schema Generation}
\label{sec:sub_schema_generation_step}
Effective in-domain Text-to-SQL synthesis demands controllability, semantic fidelity, and comprehensive schema coverage. However, considering the entire database schema when generating synthetic data is often impractical—especially for large-scale databases—due to the context length limitations of large language models (LLMs), reduced control over the generation process, and challenges in maintaining high semantic alignment between natural language questions and their corresponding SQL queries.

To address these issues, we introduce a sub-schema generation approach that partitions the original database into smaller, manageable segments. Each sub-schema is composed of a limited number of relationally connected tables and portions of their columns, enabling focused and interpretable contexts for generation. This strategy offers several key advantages. First, it provides fine-grained control over the scope of each example, allowing the generation process to focus on specific parts of the schema. Second, by narrowing the schema context, it reduces noise and improves the semantic alignment between natural language questions and their corresponding SQL queries. Third, by systematically varying the structure and composition of sub-schemas, we ensure that all schema elements appear across diverse relational settings. This not only enhances the diversity of the generated data but also ensures thorough coverage of the database, which is especially important for large and complex schemas. Overall, this step lays a strong foundation for generating high-quality, diverse, and semantically grounded Text-to-SQL pairs. The complete sub-schema generation procedure is formally described in Algorithms ~\ref{alg:sub_schema_construction_algo}, ~\ref{alg:table_level_ss_gen} and ~\ref{alg:colum_level_ss_gen}, and is further illustrated in Figure ~\ref{fig:synthetic_t2s_generation} to provide a visual understanding of the process.

\begin{algorithm}[htbp]
\caption{Sub-Schema Construction Algorithm (\texttt{ConstructSubSchemas})} \label{alg:sub_schema_construction_algo}
\begin{algorithmic}[1]
\Require window $w$, stride $s$, table counts for sub-schemas $tc$ 
\Ensure \texttt{columnLevelSubSchemas}
\State \texttt{tableLevelSubSchemas} $\gets$ \Call{GenTableLevelSubSchemas}{$tc$}
\State \texttt{columnLevelSubSchemas} $\gets$ \Call{GenColumnLevelSubSchemas}{\texttt{tableLevelSubSchemas}, $w$, $s$}
\end{algorithmic}
\end{algorithm}

\subsubsection{Table Level Sub-Schema Generation}
\label{sec:table_level_sub_schema_generation}
We begin sub-schema construction at the table level by identifying meaningful subsets of tables that preserve database integrity and relational connectivity. As outlined in Algorithm~\ref{alg:table_level_ss_gen}, all joinable tables are first extracted using the foreign key constraints defined in the database schema. In cases where foreign key information is incomplete or missing, the schema must be updated or manually annotated to restore valid join paths and ensure relational completeness. This step is essential for constructing sub-schemas capable of supporting executable and semantically meaningful SQL queries. In the next step, we enumerate all valid table combinations that maintain joinability—i.e., every table within a combination must be transitively joinable with the others. 

To ensure practical relevance and avoid overly complex sub-schemas, we introduce a hyperparameter that limits the maximum number of tables per sub-schema. This constraint reflects the observation that real-world SQL queries rarely involve all tables in a schema, especially in large databases. For instance, in the \textit{Card Games} database from the BIRD dev set, which contains 6 tables, setting the maximum number of tables per sub-schema to 3 can lead to more manageable and potentially more realistic combinations. While this value may vary depending on the schema complexity and target use case, it serves as a practical upper bound that helps constrain the sub-schema space. In addition to guiding the realism of generated queries, this hyperparameter also prevents an explosion in the number of possible sub-schemas, making the overall synthesis process more computationally feasible. To offer further insights, Appendix~\ref{appendix:ss_and_synth_data_statistics} presents sub-schema statistics across a range of SING-SQL parameters.

\begin{algorithm}[htbp]
\caption{Table Level Sub Schema Construction (Generation) Algorithm (\texttt{GenTableLevelSubSchemas})}\label{alg:table_level_ss_gen}
\begin{algorithmic}[1]
\Require table counts for sub-schemas $tc$ 
\Ensure \texttt{tableLevelSubSchemas} 
\Comment{tlss = table level sub-schemas}
\State \texttt{joinableTables} $\gets$ \Call{GetJoinableTables}{}
\State \texttt{AllTableCombinations} $\gets$ \Call{GetAllDBTableComb}{}
\State tableLevelSubSchemas $\gets$ \Call{FindSubSchemas}{\texttt{joinableTables, AllTableCombinations}} \Comment{Ensure that all tables in a table combination can joinable and are able to construct a sub-schema}
\end{algorithmic}
\end{algorithm}

\subsubsection{Column Level Sub-Schema Generation}
\label{sec:col_level_sub_schema_generation}
Building on the table-level sub-schemas, we further refine the schema context by selecting subsets of columns for each table, yielding column-level sub-schemas. For each table in a given table-level sub-schema, we begin by including all connection columns, defined as primary keys and columns participating in foreign key relationships. These columns are essential for preserving the relational structure across tables and ensuring that the synthesized SQL queries will be executable and semantically valid.

The remaining non-connection columns are then considered for selective inclusion using a controlled sliding window strategy. Two hyperparameters govern this process: the window size $w$, which specifies the number of non-connection columns selected at a time, and the stride $s$, which determines the number of positions the window shifts after each selection. Prior to windowing, the non-connection columns of each table are randomly shuffled to increase variation. The sliding window is then applied to generate multiple column subsets for each table.

These subsets are combined across tables using a Cartesian product to yield diverse column-level sub-schemas derived from the original table-level sub-schema. Compared to an exhaustive enumeration of all possible column combinations—which would lead to a combinatorial explosion in the number of sub-schemas, making exhaustive enumeration impractical—this strategy offers a scalable and tunable mechanism for achieving both coverage and diversity in the generated data.

This hierarchical construction—first over tables, and then over columns—yields a rich and diverse collection of sub-schemas with varying granularity. It enables fine-grained control over schema exposure during data generation while maintaining syntactic validity and semantic coherence. A visual overview of this process is provided in Figure~\ref{fig:synthetic_t2s_generation}, and the full column-level sub-schema construction procedure is detailed in Algorithm~\ref{alg:colum_level_ss_gen}.In Appendix~\ref{appendix:ss_and_synth_data_statistics}, we present sub-schema statistics for various SING-SQL parameters to provide further insights.

\begin{algorithm}[htbp]
\caption{Column-Level Sub-Schema Construction (\texttt{GenColumnLevelSubSchemas})}
\label{alg:colum_level_ss_gen}
\begin{algorithmic}[1]
\Require window size $w$, stride $s$, \texttt{tableLevelSubSchemas}
\Ensure \texttt{columnLevelSubSchemas}

\State Initialize \texttt{columnLevelSubSchemas} $\gets$ [ ]
\For{each \texttt{tlss} in \texttt{tableLevelSubSchemas}}

    \State Initialize \texttt{tableColumnParts} $\gets$ empty dictionary

    \For{each \texttt{table} in \texttt{tlss}}

        \State \texttt{connCols} $\gets$ \Call{GetConnectionColumns}{\texttt{table}} 
        \State \texttt{nonConnCols} $\gets$ \texttt{AllColumns(table)} $\setminus$ \texttt{connCols}

        \If{\texttt{nonConnCols} is empty}
            \State \texttt{tableColumnParts[table]} $\gets$ [\texttt{connCols}]
        \Else
            \State Randomly shuffle \texttt{nonConnCols}
            \State Initialize \texttt{colParts} $\gets$ [ ]
            \State $i \gets 0$
            \While{$i < $ \texttt{len(nonConnCols)}}
                \State \texttt{portion} $\gets$ \texttt{nonConnCols[$i$ : $i + w$]}
                \State \texttt{portion} $\gets$ \texttt{connCols} $+$ \texttt{portion}
                \State Append \texttt{portion} to \texttt{colParts}
                \State $i \gets i + s$
            \EndWhile
            \State \texttt{tableColumnParts[table]} $\gets$ \texttt{colParts}
        \EndIf

    \EndFor

    \State \texttt{colLevelSchemas} $\gets$ Cartesian product of all table-column partitions in \texttt{tableColumnParts}

    \For{each \texttt{schema} in \texttt{colLevelSchemas}}
        \State Append \texttt{newSchema} to \texttt{columnLevelSubSchemas}
    \EndFor

\EndFor
\State \Return \texttt{columnLevelSubSchemas}
\end{algorithmic}
\end{algorithm}

\subsection{Synthetic Text-to-SQL Generation}
\label{sec:synthehic_t2s_generation_step}
Following the construction of diverse sub-schemas, the second stage of our framework focuses on synthesizing high-quality natural language and SQL query pairs. Unlike prior work~\cite{li-2024-codes, yang-etal-2024-synthesizing-t2s-data-from-weak-and-strong-llms} that typically follows a Text-to-SQL approach—generating natural language questions first and then deriving the corresponding SQL—we adopt the reverse paradigm. Specifically, we first generate SQL queries and subsequently translate them into natural language questions. This decision is motivated by findings from prior work such as OMNI-SQL~\cite{li2025omnisqlsynthesizinghighqualitytexttosql}, which demonstrate that SQL-to-Text generation is generally more reliable due to the greater flexibility and expressiveness of natural language.

For each sub-schema, the large language model (LLM) is prompted to generate $N$ SQL queries spanning four predefined complexity levels: simple, moderate, challenging, and window. The first three levels are consistent with those adopted in established benchmarks such as BIRD, while the window level is introduced to explicitly control for the inclusion of window functions—an area where LLMs tend to struggle when left unguided. By defining this complexity level explicitly, we avoid relying solely on model inference to ensure such queries are included. In our work, we use $N=3$, resulting twelve SQL-Text pair generation for each sub-schema. To encourage broad schema coverage, we explicitly instruct the model to utilize all available tables and columns in the sub-schema during SQL generation. 

Although SQL-to-Text synthesis yields more semantically aligned pairs compared to the reverse direction, errors may still occur—such as flawed natural language translations or illogical SQL-question pairs (see Appendix~\ref{appendix:flawed_sql_to_text_translation} and Appendix~\ref{appendix:illogical_sql_to_text_pairs}).  To identify and eliminate such issues, we introduce a validation stage in which the LLM is employed as a judge (LLM-as-a-judge). Each SQL–Text pair is evaluated within the context of its sub-schema, and pairs exhibiting misalignment, ambiguity, or logical flaws are discarded.

Subsequently, we assess the executability of each synthetic SQL query. Queries that fail to execute are passed through an automated repair step using the LLM. If the query remains non-executable after correction attempts, it is excluded from the dataset. Once a validated and executable set of SQL–Text pairs is obtained, we further enrich the dataset by generating accompanying reasoning traces. These traces are constructed using a divide-and-conquer prompting strategy inspired by CHASE-SQL~\cite{pourreza2025chasesql}, which aims to improve the interpretability of the data and instructional signal of the synthetic examples for LLMs.

To promote sufficient schema coverage in the final dataset, we introduce an additional column-focused synthesis step. Following the first data generation round, we calculate the occurrence frequency of each column within the synthesized SQL queries. Columns that appear fewer times than a predefined threshold are considered underrepresented and are flagged for targeted inclusion. Sub-schemas containing these low-frequency columns are then reselected, and a second round of SQL–Text generation is performed. During this iteration, the LLM is explicitly instructed to focus on incorporating the underused columns into the generated SQL queries. While this strategy does not guarantee equal representation across all columns, it effectively eliminates extremely low-frequency cases and enforces a minimum level of exposure for each column, thereby improving overall schema coverage.

An overview of the complete synthesis workflow is illustrated in Figure~\ref{fig:synthetic_t2s_generation}, and the corresponding procedure is detailed in Algorithm~\ref{alg:GenT2S}.

\begin{algorithm}[htbp]
\caption{Synthetic Text-to-SQL Generation Algorithm (\texttt{GenT2S})} \label{alg:GenT2S}
\begin{algorithmic}[1]
\Require \texttt{columnLevelSubSchemas}
\Ensure \texttt{t2sExamples}
\State Initialize  \texttt{t2sExamples} $\gets$ []
\For{each \texttt{subSchema} in \texttt{columnLevelSubSchemas}}
    \State SQL2TextExamples $\gets$ \Call{GenerateWithLLM}{}
    \For{each \texttt{SQL2TextItem} in \texttt{SQL2TextExamples}}
        \State isLogical $\gets$ \Call{EvaluateSQL2TextItem}{SQL2TextItem} 
        \If{isLogical}
            \State \texttt{isExecutable} $\gets$ \Call{ExecuteSQL}{\texttt{SQL}}
            \If{isExecutable is False}
                \State \texttt{SQL2TextItem} $\gets$ \Call{FixSQL}{\texttt{SQL2TextItem}}
            \EndIf
            \State \Call{GenerateReasoning}{\texttt{SQL2TextItem}}
        \State Append \texttt{SQL2TextItem} to  \texttt{t2sExamples}
        \EndIf
    \EndFor
\EndFor
\State \Return \texttt{t2sExamples}
\end{algorithmic}
\end{algorithm}

\section{Data Statistics}
\label{sec:data_statistics}

To better understand the characteristics of our synthetic dataset and to compare it with the BIRD benchmark, we analyze both dataset composition and SQL query statistics across multiple dimensions. In particular, we examine question distributions across levels, join counts, and aggregation usage, which are widely adopted indicators of SQL complexity and diversity. We also investigate schema coverage, as incomplete coverage may hinder reliable evaluation.

Table~\ref{tab:question_counts_per_level} summarizes the distribution of questions across levels for the \textit{California Schools} database. BIRD-Dev is dominated by simple and moderate queries (94.4\% in total), with very few challenging (5) or window (2) queries. In contrast, our synthetic splits yield a substantially larger number of questions with a more balanced distribution, including large number of challenging and window queries. This broader coverage ensures that models trained on the synthetic dataset are exposed to a wider variety of SQL patterns. This also enables evaluation under more realistic data analysis scenarios, where complex queries involving multiple tables and conditions are prevalent.

\begin{figure}[htbp]
  \centering
  \includegraphics[width=0.8\linewidth]{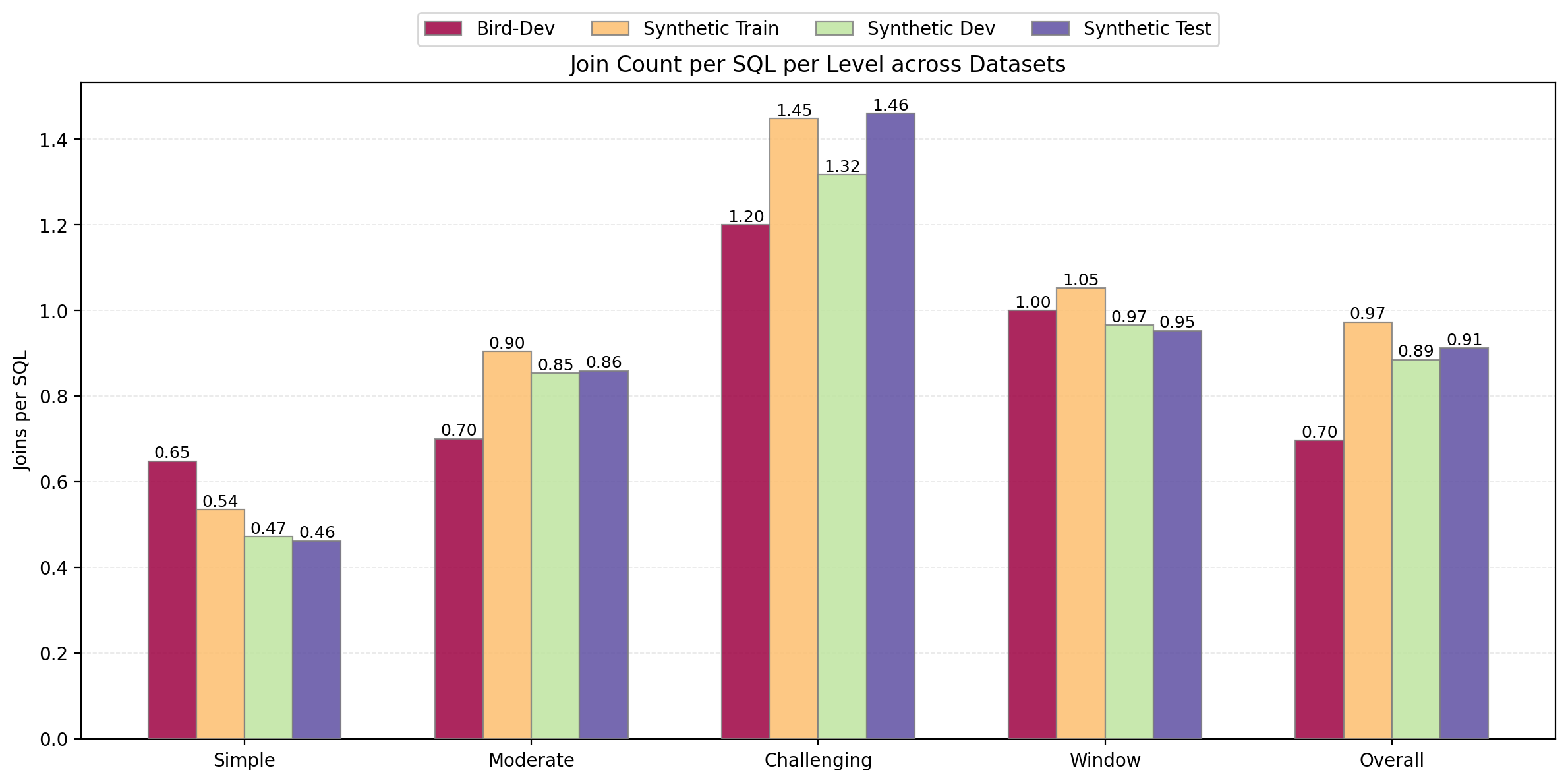}
  \caption{Join Count Comparison of Bird and Synthetic Data}
  \label{fig:join_cnt_comparison}
\end{figure}

Figure~\ref{fig:join_cnt_comparison} reports the average number of joins per SQL across different difficulty levels. We observe that the synthetic data exhibits comparable or slightly higher join complexity than BIRD, except simple questions. This suggests that our generation process successfully incorporates complex relational reasoning patterns. 

\begin{figure}[htbp]
  \centering
  \includegraphics[width=0.8\linewidth]{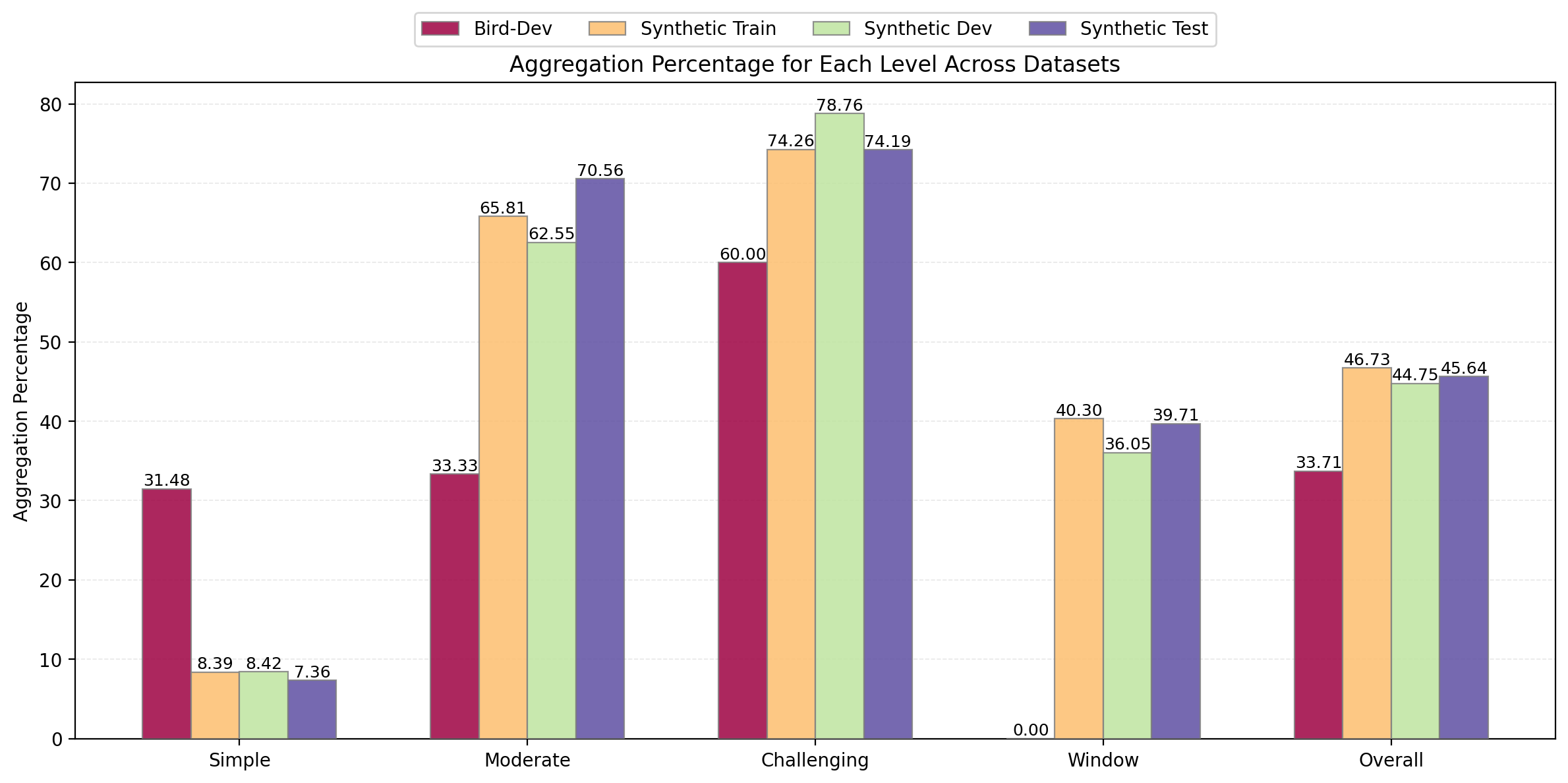}
  \caption{Aggregation Comparison of Bird and Synthetic Data}
  \label{fig:aggregation_comparison}
\end{figure}

Similarly, Figure~\ref{fig:aggregation_comparison} illustrates the distribution of queries with aggregation operators. While BIRD-Dev has a larger fraction of aggregations in the simple category (31.48\%), our synthetic data introduces a higher proportion of aggregation in moderate, challenging, and window queries, reaching up to 78.76\% in the challenging level.

Table~\ref{tab:unused_cols_california_schools} further highlights a limitation of the BIRD development split: 15 columns (16.85\% of the schema) in the \textit{California Schools} database remain completely unused. This under-utilization implies that BIRD-Dev cannot fully measure the performance of Text-to-SQL systems on queries involving these columns, nor capture potential interactions among them. In contrast, our synthetic dataset achieves complete coverage across all columns in the train, dev, and test splits, thereby enabling more comprehensive evaluation and supervision over the entire schema. Column coverage comparison for California Schools database provided in Appendix~\ref{appendix:column_coverage_comparison_california_schools}.

Together, these results demonstrate that the synthetic dataset is not only well aligned with the BIRD benchmark but also offers richer coverage of SQL constructs and schema elements at higher complexity levels, providing stronger supervision for training and evaluation of Text-to-SQL systems. Moreover, since our SING-SQL framework is database-agnostic, it can be applied to arbitrary databases, facilitating better schema alignment for LLMs and enabling more comprehensive evaluation.

\begin{table}[htbp]
\centering
\scriptsize
\caption{Question counts per difficulty level for the \textit{California Schools} database in BIRD-Dev and synthetic splits.}
\label{tab:question_counts_per_level}
\resizebox{\columnwidth}{!}{%
\begin{tabular}{lccccc}
\toprule
Dataset & Overall & Simple & Moderate & Challenging & Window \\
\midrule
\texttt{BIRD-Dev} & 89 & 54 & 30 & 5 & 2 \\
\texttt{Synthetic Train} & 34,266 & 8,685 & 8,556 & 8,046 & 9,286 \\
\texttt{Synthetic Dev} & 1,124 & 297 & 259 & 259 & 319 \\
\texttt{Synthetic Test} & 1,124 & 299 & 248 & 248 & 340 \\
\bottomrule
\end{tabular}%
}
\end{table}

\begin{table}[htbp]
\centering
\scriptsize
\caption{Unused column statistics for the \textit{California Schools} database.}
\label{tab:unused_cols_california_schools}
\resizebox{\columnwidth}{!}{%
\begin{tabular}{lcc}
\toprule
Dataset  & Unused Column Count & Unused Column Rate (\%) \\
\midrule
\texttt{Bird-Dev}  & 15 & 16.85 \\
\texttt{Synthetic Train} & 0 & 0.0 \\
\texttt{Synthetic Dev} & 0 & 0.0 \\
\texttt{Synthetic Test} & 0 & 0.0 \\
\bottomrule
\end{tabular}%
}
\end{table}

\begin{figure*}[htbp]
  \centering
  \includegraphics[width=0.8\linewidth]{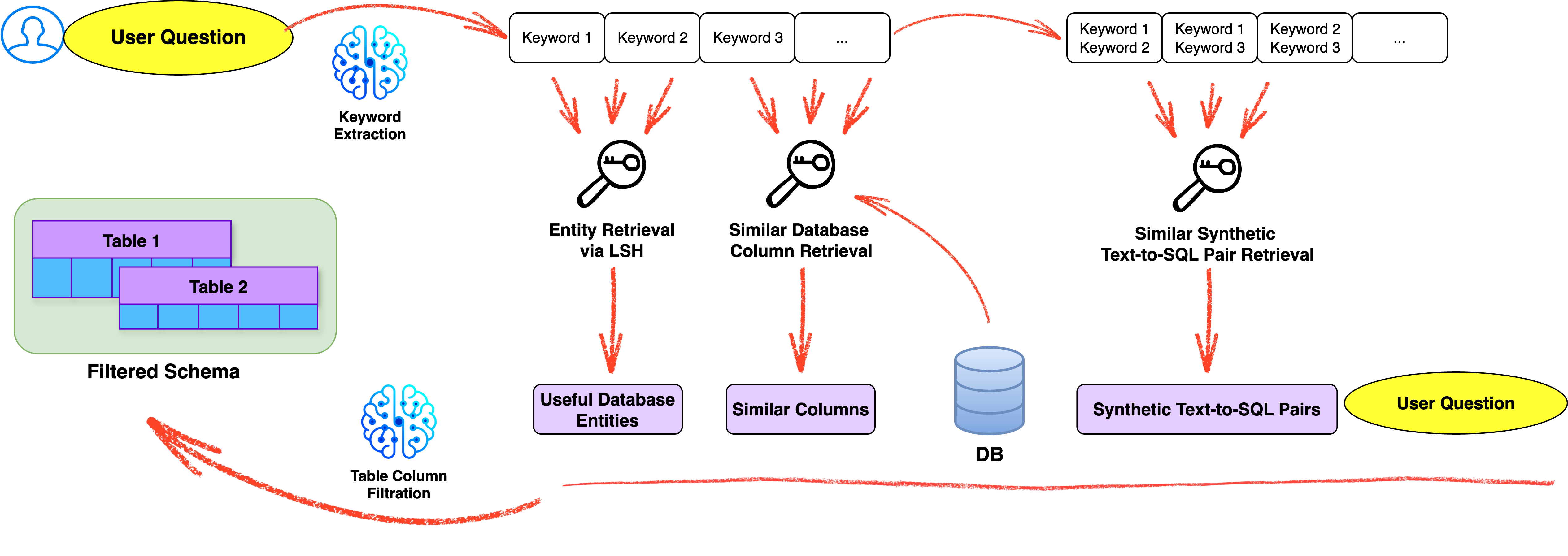}
  \caption{Overview of the SING-SQL Schema Filtering}
  \label{fig:sing_sql_schema_filtering}
\end{figure*}

\section{Text-to-SQL Translation}
\label{sec:text_to_sql_translation_details}
A typical Text-to-SQL translation pipeline consists of four core components: schema linking, candidate generation, candidate refinement, and candidate selection. In this work, our focus is on evaluating the SQL generation capabilities of models without constructing a full agentic pipeline. Therefore, we restrict our experiments to the schema linking and candidate generation stages. As detailed in Section~\ref{sec:related_work}, unlike prior studies, our approach leverages synthetically generated \emph{in-domain} Text-to-SQL data for both schema linking and the construction of few-shot examples for SQL generation.

\subsection{Schema Linking}
\label{sec:t2s_translation_schema_linking}

Our schema linking process is partly driven by synthetically generated in-domain Text-to-SQL pairs. To retrieve relevant synthetic examples and associated database items, we first extract keywords from the input question following the approach in CHESS~\cite{talaei2024chess}. Relying solely on individual keywords can be limiting, as questions often contain multiple terms whose combinational usage influences retrieval quality. To address this, we expand the search space by forming \emph{keyword pairs}, enabling more precise matching against synthetic examples.

For retrieving synthetic question–SQL pairs, we implement a retrieval process that operates over both the question and SQL text. Specifically, we experiment with two methods: (1) BM25 lexical matching, and (2) semantic search using dense vector representations. While both approaches are capable of identifying relevant examples, semantic search is substantially slower in practice. Considering real-world deployment scenarios, we therefore adopt BM25 as our primary retrieval method due to its efficiency. In parallel, we retrieve relevant database columns and entities, following CHESS~\cite{talaei2024chess}, to further enhance the schema linking process. Finally, we filter the candidate database tables using the retrieved synthetic examples, relevant columns, and entities, ensuring a focused and schema-consistent context for downstream SQL generation.

\subsection{Candidate Generation}
After schema filtering, we generate multiple SQL candidates following the general strategy adopted in prior work ~\cite{wang-2024-macsql, luo2024SuperSQL, lee-2024-mcs-sql, talaei2024chess, pourreza2025chasesql, XiYanSQL, sheng2025cscsqlcorrectiveselfconsistencytexttosql, sheng2025slmsqlexplorationsmalllanguage}. Here again, our in-domain synthetic dataset plays a central role. Using the same keyword-pair retrieval process described in the schema linking stage, we identify the most relevant synthetic examples for the input question. We then rank these examples by similarity and select the top-$k$ as few-shot demonstrations for the LLM. This targeted retrieval ensures that the few-shots are both semantically relevant and schema-consistent. In addition, we incorporate detailed column metadata—such as semantic descriptions or representative values—into the prompt, providing richer context for the model during SQL generation.

\section{Experiment Settings}
\label{sec:experiment_settings}

\subsection{Dataset}
\label{sec:dataset}
We conduct experiments on the California Schools database from the BIRD development benchmark~\cite{li-2024-bird-sql}, as well as on our synthetically generated in-domain dataset produced by the database-agnostic SING-SQL framework. While BIRD-Dev provides a limited set of 89 questions, heavily skewed toward simple and moderate difficulty, our synthetic dataset achieves full schema coverage and a more balanced distribution across difficulty levels, including a substantial number of challenging and window queries, as shown in Table~\ref{tab:question_counts_per_level}.

\subsection{Evaluation Metrics}
\label{sec:evaluation_metrics}
We evaluate model performance using two official metrics from the BIRD~\cite{li-2024-bird-sql} benchmark: Execution Accuracy (EX) and Soft F1-score. Execution Accuracy (EX) measures whether the execution results of predicted SQL queries exactly match those of the gold queries. While EX accounts for the semantic equivalence of structurally different SQLs, it remains a strict metric: a single discrepancy, such as column order variation, leads to a score of zero even if the query produces an otherwise correct result. This rigidity makes EX useful for measuring exact correctness but less suitable for reflecting practical robustness.

To address these limitations, the BIRD benchmark introduced the Soft F1-score, which compares the overlap between predicted and gold result tables in a precision–recall framework. By tolerating column reordering and minor inconsistencies such as missing values, Soft F1 provides a more faithful estimate of how well a model captures the intended semantics of a query. For example, when a predicted query retrieves the correct tuples but orders or partially aligns columns differently, Soft F1 still rewards the overlap instead of reducing the score to zero.

In this work, we primarily rely on the Soft F1-score for evaluation. We argue that Soft F1 better reflects the reliability of Text-to-SQL systems in real-world deployments, where users care about retrieving semantically correct answers rather than perfectly ordered result tables. Execution Accuracy is still reported for completeness and comparability with prior work, but we consider Soft F1 to be a more informative measure of model performance and a stronger indicator of progress in in-domain Text-to-SQL translation.

\subsection{Models and Hyperparameters}
\label{sec:models_and_hyperparameters}
For synthetic data generation, we employ the Gemini-2.5-Flash model across all stages of the SING-SQL framework, ensuring efficient and consistent synthesis of SQL--Text pairs with high semantic fidelity. Again, we employ Gemini-2.5-Flash in LLM-based schema filtering for accurate and comparable filtering performance. For downstream Text-to-SQL translation, we adopt Qwen2.5-Coder-1.5B Instruct and Qwen2.5-Coder-3B Instruct~\cite{hui2024qwen25codertechnicalreport} as the base model. 
To adapt this backbone to in-domain settings, we apply supervised fine-tuning (SFT), yielding specialized variants that we refer to as SingSQL-LM. 

The fine-tuning process is carried out using parameter-efficient Low-Rank Adaptation (LoRA)~\cite{hu_2022_lora_lowrankadaptation}, explored under two configurations: LoRA rank 32 and LoRA rank 64. In the rank-32 setting, we configure LoRA Alpha as 32, a learning rate of $1.0 \times 10^{-4}$, an effective batch size of 8, a warm-up ratio of 0.1, and cosine scheduling. In the rank-64 setting, we set LoRA Alpha as 64, a learning rate of $7.5 \times 10^{-5}$, an effective batch size of 8, a warm-up ratio of 0.1, and the same cosine scheduler. 
All models are trained for two epochs. Fine-tuning is performed using the \texttt{Unsloth}\textsuperscript{1} library to enable efficient large-scale adaptation. 

All experiments are conducted on a single NVIDIA A40 GPU equipped with 48 GB of VRAM.

\section{Results}
\label{sec:results}
In Section~\ref{t2s_translation_performance}, we report the performance of the fine-tuned \textsc{SingSQL-LM} models on Bird benchmark \cite{li-2024-bird-sql} and synthetic datasets. Section~\ref{schema_filtering_performance} presents an evaluation of schema filtering techniques that leverage the generated synthetic data. Section~\ref{context_management_performance} examines the effect of different context management strategies on model performance.

\subsection{Text-to-SQL Translation Performance}
\label{t2s_translation_performance}

\begin{table*}[htbp]
\centering
\caption{Comparison of system performance on the California Schools subset of the BIRD development benchmark.}
\label{tab:comparison_our_lm_with_others_on_bird_california_schools}

\begin{adjustbox}{max width=\linewidth}
\begin{tabular}{lcccccccccccc}
\toprule
\multirow{2}{*}{\textbf{System}} & \multicolumn{4}{c}{\textbf{Candidate Count = 8}} & \multicolumn{4}{c}{\textbf{Candidate Count = 16}} & \multicolumn{4}{c}{\textbf{Candidate Count = 32}} \\
\cmidrule(lr){2-5} \cmidrule(lr){6-9} \cmidrule(lr){10-13}
& \textbf{EX UB} & \textbf{EX LB} & \textbf{F1 UB} & \textbf{F1 LB} & \textbf{EX UB} & \textbf{EX LB} & \textbf{F1 UB} & \textbf{F1 LB} & \textbf{EX UB} & \textbf{EX LB} & \textbf{F1 UB} & \textbf{F1 LB}  \\
\midrule

CODES-1B  & 43.82 & 6.74 & 49.40 & 9.32 & 44.94 & 3.37 & 51.90 & 5.02 & 50.56 & 1.12 & 60.72 & 44.94 \\
SLM-SQL-1.5B \textsuperscript{\dag} & 40.45 & 0.0 & 47.08 & 0.0 & 52.81 & 0.0 & 59.84 & 0.0 & 59.55 & 0.0 & 64.52 & 0.0 \\
CODES-3B  & 50.56 & 11.24 & 56.46  & 14.50 & 56.18 & 3.37 & 62.17 & 5.80 & 60.67 & 1.12 & 66.66 & 1.77  \\
OmniSQL-7B \textsuperscript{\ddag} & 73.03 & 35.96 & 80.36 & 44.81 & -- & -- & -- & -- & -- & -- & -- & -- \\
CscSQL-Grpo-Qwen2.5-Coder-7B-Instruct & 66.29 & 1.12 & 72.02 & 1.12 & 74.15 & 0.0 & 76.99 & 0.0 & 77.55 & 0.0 & 81.45 & 0.0 \\
CHESS & 61.80 & 23.60 & 73.04 & 32.83 & 69.66 & 14.61 & 75.73 & 18.32 & 76.40 & 11.24 & 82.63 & 16.13 \\

\midrule
\textbf{SingSQL-LM-1.5B-R32} & 41.57 & 0.0 & 53.26 & 0.0 & 52.80 & 0.0 & 64.75 & 0.0 & 59.55 & 0.0 & 67.55 & 0.0 \\
\textbf{SingSQL-LM-1.5B-R64} & 53.93 & 0.0 & 63.89 & 0.0 & 58.43 & 0.0 & 67.20 & 0.0 & 64.04 & 0.0 & 73.82 & 0.0 \\
\textbf{SingSQL-LM-3B-R32} & 58.42 & 0.0 & 71.06 & 1.00 & 67.42 & 0.0 & 77.61  & 0.56 & 70.79 & 0.0 & 81.76 & 0.0  \\
\textbf{SingSQL-LM-3B-R64} & 64.04 & 1.12 & 73.61 & 6.26 & 68.54 & 0.0 & 79.21 & 0.85 & 73.03 & 0.0 & 82.87 & 0.0  \\
\bottomrule
\end{tabular}
\end{adjustbox}
\parbox{\linewidth}{\scriptsize\textbf{R} represents the LoRA rank. \textbf{UB} represents upper bound and \textbf{LB} represents lower bound.  We do not provide 3B models of CSC-SQL as we found its output inconsistent.\textsuperscript{\dag} Only the SQL generator model is used for SLM-SQL. \textsuperscript{\ddag} The values are calculated from candidate SQL queries directly given by the OmniSQL authors in Github\textsuperscript{2}.}
\end{table*}
\footnotetext[1]{\url{https://unsloth.ai}}
\footnotetext[2]{\url{https://github.com/RUCKBReasoning/OmniSQL/tree/main/OmniSQL_prediction_results}}

\begin{table*}[htbp]
\centering
\caption{Comparison of system performance (Candidate SQL count is 8) on the Synthetic California Schools Development and Test Sets of SING Framework. }
\label{tab:comparison_our_lm_with_others_on_synthetic_california_schools}
\begin{adjustbox}{max width=\linewidth}
\small
\begin{tabular}{lcccccccc}
\toprule
\multirow{2}{*}{\textbf{System}} & \multicolumn{4}{c}{\textbf{SING-Dev}} & \multicolumn{4}{c}{\textbf{SING-Test}} \\
\cmidrule(lr){2-5} \cmidrule(lr){6-9}
& \textbf{EX UB} & \textbf{EX LB} & \textbf{F1 UB} & \textbf{F1 LB} & \textbf{EX UB} & \textbf{EX LB} & \textbf{F1 UB} & \textbf{F1 LB} \\
\midrule

CODES-1B  & 17.97 & 0.53 & 28.70 & 1.96 & 16.73 & 0.36 & 27.11 & 1.79 \\
SLM-SQL-1.5B & 24.02 & 0.44 & 30.24 & 0.69 & 22.15 & 0.27 & 27.89 & 0.43 \\
CODES-3B & 22.42 & 0.53 & 34.57 & 2.18 & 20.20 & 0.71 & 33.07 & 2.80 \\
CscSQL-Grpo-Qwen2.5-Coder-7B-Instruct & 30.60 & 0.80 & 39.61 & 1.04 & 27.85 & 0.36 & 37.29 & 0.59 \\
CHESS & 38.70 & 10.23 & 53.44 & 16.79 & 40.75 & 9.16 & 55.24 & 18.24 \\
\midrule
\textbf{SingSQL-LM-1.5B-R32} & 55.43 & 1.69 & 65.90 & 3.38 & 55.07 & 0.98 & 65.27 & 2.28 \\
\textbf{SingSQL-LM-1.5B-R64} & 56.85 & 2.22 & 68.05 & 4.04 & 58.10 & 1.60 & 66.59 & 3.00 \\
\textbf{SingSQL-LM-3B-R32} & 62.99 & 3.73 & 73.14 & 8.95 & 63.52 & 3.83 & 71.74 & 7.84 \\
\textbf{SingSQL-LM-3B-R64} & 65.21 & 5.25 & 75.33 & 10.58 & 64.08 & 5.16 & 72.37 & 9.87 \\
\bottomrule
\end{tabular}
\end{adjustbox}
\end{table*}

In evaluating translation performance, we restrict our analysis to SQL generation stage where filtered schema used, deliberately excluding candidate selection mechanisms. This ensures that results reflect the intrinsic quality of the generated SQL queries, rather than the effectiveness of downstream ranking or selection modules. We report both Execution Accuracy (EX) and Soft F1-score ~\cite{li-2024-bird-sql}, under lower- and upper-bound settings, across varying candidate counts. Results are presented on the California Schools subset of the BIRD development benchmark and on the synthetic California Schools data generated with the SING framework in Table~\ref{tab:comparison_our_lm_with_others_on_bird_california_schools}  and Table~\ref{tab:comparison_our_lm_with_others_on_synthetic_california_schools}, respectively.

On BIRD-Dev, shown in Table~\ref{tab:comparison_our_lm_with_others_on_bird_california_schools}, SingSQL-LM consistently outperforms comparable baselines. Among 1.5B-scale models, SingSQL-LM deliver substantial gains, reaching up to 73.82\% Soft F1 UB and 64.04\% EX UB with 32 candidates—9.30\% and 4.49\% gain over SLM-SQL-1.5B (64.52\% Soft F1 UB, 59.55\% EX UB) in terms of Soft F1 and EX respectively and clearly surpassing CODES-1B and 3B. At the 3B scale, SingSQL-LM achieves state-of-the-art performance among open models. With 32 candidates, SingSQL-LM-3B-R64 obtains 82.87\% Soft F1 UB and 73.03\% EX UB, outperforming the best 3B-scale baseline by +16.21 in Soft F1 and +12.36 in EX. Remarkably, SingSQL-LM-3B-R64 surpasses the 7B-scale CscSQL model, demonstrating that compact models can achieve superior performance with targeted in-domain supervision. Lower-bound scores remain close to zero across most settings due to the absence of refinement or self-consistency, but the strong upper bounds demonstrate that the models reliably generate high-quality SQL candidates.


On the synthetic evaluation splits, given in Table~\ref{tab:comparison_our_lm_with_others_on_synthetic_california_schools}, the advantages of SingSQL-LM become even more pronounced. SingSQL-LM-1.5B-R64 achieves 68.05\% Soft F1 UB and 56.85\% EX UB on SING-Dev, and 66.59\% Soft F1 UB and 58.10\% EX UB on SING-Test, outperforming larger baselines such as CscSQL-Grpo-7B-Instruct, which remain below 40\% Soft F1 and 31\% EX. The 3B-scale variants further improve these results, with SingSQL-LM-3B-R64 reaching 75.33\% Soft F1 UB and 65.21\% EX UB on SING-Dev, and 72.37\% Soft F1 UB and 64.08\% EX UB on SING-Test. These results exceed all reported baselines by wide margins, underscoring that fine-tuning on SING-SQL’s high-coverage synthetic data enables robust in-domain generalization and offers a reliable evaluation path for database-specific performance.


Together, these findings emphasize two main points. First, SingSQL-LM demonstrates state-of-the-art performance among open models, while rivaling or surpassing 7B-scale systems. Second, the substantial improvements observed on the synthetic development and test sets underscore the value of database-specific supervision: models fine-tuned with SING-SQL not only outperform cross-domain baselines but also demonstrate consistent robustness across synthetic evaluation splits. This confirms the utility of synthetic in-domain data as a scalable path toward enterprise-grade Text-to-SQL systems.

\subsection{Schema Filtering Performance}
\label{schema_filtering_performance}

Table~\ref{tab:schema_filtering_performance} reports schema filtering results on the California Schools database from the BIRD development split, comparing multiple filtering strategies that leverage synthetic Text-to-SQL pairs. We evaluate recall and precision at both the table and column levels, along with strict schema recall rate (SRR)~\cite{cao_2024_rsl_sql}. While both precision and recall are informative, recall is particularly critical in Text-to-SQL systems~\cite{is_long_context_all_you_need}, as omitting any part of the ground-truth schema inevitably leads to incorrect translations.  

As described in Section~\ref{sec:t2s_translation_schema_linking}, our schema linking process relies on synthetically generated Text-to-SQL pairs. We evaluate schema filtering performance under different retrieval methods. Since our approach does not filter tables, table recall remains consistently at 100\%. For each keyword pair, we extract a single synthetic Text-to-SQL example. When all retrieved examples are considered, column recall reaches 87.23\% with BM25 and 94.20\% with vector-based retrieval. However, augmenting prompts with all retrieved examples is impractical for downstream tasks due to increased inference time and cost. To this end, we restrict the context to the top-6 most relevant examples, which reduces column recall to 71.93\% (BM25) and 70.53\% (vector retrieval). These results indicate that relying solely on synthetic examples retrieved via basic similarity methods yields limited effectiveness and can negatively impact overall Text-to-SQL translation.  

To mitigate this, we incorporate an LLM-based filtering stage on top of retrieval. This hybrid strategy significantly improves the recall with the limited number of example. Specifically, BM25-Top6+LLM achieves 97.45\% column recall with 46.77\% precision, corresponding to an SRR of 88.76. Similarly, Vec-Top6+LLM reaches 97.91\% recall with 47.52\% precision, yielding the best SRR of 91.01. Since all tables are always included in the filtered schema along with their primary and foreign key columns, our precision scores are lower compared to systems that apply stricter pruning. The performance of schema filtering across different mechanisms on synthetic datasets is further detailed in Appendix~\ref{appendix:schema_filtering_performance_on_synthetic_data}.  

\begin{table}[htbp]
\centering
\caption{Schema Filtering Performance on California Schools database in Bird Development Split.}
\label{tab:schema_filtering_performance}
\begin{threeparttable}
\begin{tabular}{lccccc}
\toprule
\textbf{Method} & \textbf{TR} & \textbf{TP} & \textbf{CR} & \textbf{CP}  & \textbf{SRR} \\
\midrule
RSL-SQL\textsuperscript{$\diamondsuit$} & -- & -- & -- & -- & 91.80 \\
CHESS\textsuperscript{$\diamondsuit$}   & 97.69 & 89.72 & 97.12 & 69.43 & 89.70 \\
MCS\textsuperscript{$\diamondsuit$}  & -- & -- & 89.80 & -- & -- \\
\midrule
BM25-Top6              & 100 & 58.82 & 71.93 & 27.33 & 25.84 \\
Vec-Top6               & 100 & 59.29 & 70.53 & 28.90 & 24.72 \\
BM25-All               & 100 & 56.18 & 87.23 & 14.86 & 51.68 \\
Vec-All                & 100 & 56.39 & 94.20 & 12.59 & 73.03 \\
BM25-Top6 + LLM   & 100 & 58.37 & \textbf{97.45} & 46.77 & 88.76 \\
Vec-Top6 + LLM     & 100 & 57.03 & \textbf{97.91} & 47.52 & 91.01 \\
\bottomrule
\end{tabular}
\parbox{\linewidth}{\footnotesize \textsuperscript{$\diamondsuit$} Given metrics are for the whole development set of BIRD benchmark.}
\parbox{\linewidth}{\footnotesize All our methods retrieve few-shots using \emph{user-question keyword pairs}. ``BM25'' retrieves examples using BM25 algorithm; ``Vec'' retrieves examples using semantic similarity over a vector database. ``All'' uses all retrieved few-shots to construct filtered schema while ``Top6'' uses the top-6 most similar examples to the user question. ``+LLM'' applies LLM-based table column filtering leveraging retrieved synthetic examples. ``TR'' and ``TP'' represent Table Recall and Precision respectively. Similarly, ``CR'' and ``CP'' represent Column Recall and Precision respectively. ``SRR'' denotes strict schema recall rate ~\cite{cao_2024_rsl_sql}.}
\end{threeparttable}
\end{table}

Overall, these results highlight that combining synthetic in-domain supervision with the LLM-based filtering produces robust schema linking. By coupling efficient retrieval with targeted LLM refinement, SING-SQL achieves strong schema coverage while ensuring practical scalability, thus establishing a reliable foundation for downstream SQL generation.

\subsection{Context Management}
\label{context_management_performance}

We evaluate the effect of different contextual signals on SQL generation, focusing on schema, few-shot demonstrations and few-shot reasoning traces. 

For the base model, schema grounding provides the most stable supervision. As shown in Table~\ref{tab:performance_comparison_of_models_with_various_contexts}, supplying six few-shots with reasoning consistently reduces performance, while providing few-shots without reasoning improves the performance slightly, suggesting that few-shots can offer marginal benefits but reasoning traces can introduce noise.

When fine-tuned models are considered, schema-only inference emerges as the most effective configuration across all training regimes. Adding few-shots at inference consistently lowers accuracy, particularly when reasoning traces are included as seen in Table~\ref{tab:performance_comparison_of_models_with_various_contexts}. An important observation is that training with schema-aware data (T2SWS) leads to intermediate performance, whereas models fine-tuned on schema-free data (T2S) achieve the highest scores under schema-only inference. This indicates that schema-free fine-tuning allows the model to internalize SQL patterns and leverage schema more effectively during inference, while explicit schema supervision during training provides less robust transfer.

Overall, three findings stand out. First, schema grounding is the most robust contextual signal, outperforming configurations with additional few-shots for fine-tuned models. Second, schema-free fine-tuning combined with schema-only inference yields the strongest generalization to in-domain queries. Third, few-shots especially with reasoning traces tend to destabilize performance, limiting their utility once schema context is explicitly provided. Additional experiments with other fine-tuned models confirm these patterns, as detailed in Appendix~\ref{appendix:context_management_study}.

\begin{table}[htbp]
\centering
\caption{Comparison of model performance across different training and inference contexts (candidate SQL count = 8). }
\label{tab:performance_comparison_of_models_with_various_contexts}
\renewcommand{\arraystretch}{1.00} 
\begin{adjustbox}{max width=\columnwidth, scale=1} 
\begin{tabular}{ccccccccc}
\toprule
\multicolumn{3}{c}{\textbf{Training Context}} &  
\multicolumn{3}{c}{\textbf{Inference Context}} &  
\multicolumn{2}{c}{\textbf{Performance}} \\
\cmidrule(lr){1-3} \cmidrule(lr){4-6} \cmidrule(lr){7-8}
\textbf{Dataset} & \textbf{FS-C} & \textbf{FS-R} & \textbf{Schema} &
\textbf{FS-C} & \textbf{FS-R} & 
\textbf{EX UB} & \textbf{F1 UB} \\
\midrule
\multicolumn{3}{c}{Base Model} & \checkmark & 6 & \checkmark & 38.20 & 43.20 \\
\multicolumn{3}{c}{Base Model} & \checkmark & 6 & \checkmark & 47.19 & 54.94 \\
\multicolumn{3}{c}{Base Model} & \checkmark & 0 & NA & 46.06 & 54.37 \\
\midrule
T2SWS & 6 & $\times$ & \checkmark & 6 & \checkmark & 40.44 & 52.55 \\
T2SWS & 6 & $\times$ & \checkmark & 6 & $\times$ & 33.70 & 46.38 \\
T2SWS & 6 & $\times$ & \checkmark & 0 & NA & 46.07 & 56.00 \\
\midrule
T2SWS & 0 & NA & \checkmark & 6 & \checkmark & 35.96 & 49.87 \\
T2SWS & 0 & NA & \checkmark & 6 & $\times$ & 40.45 & 53.29 \\
T2SWS & 0 & NA & \checkmark & 0 & NA & 51.69 & 63.20 \\
\midrule
T2SWS, T2S & 6 & $\times$ & \checkmark & 6 & \checkmark & 30.34 & 43.49 \\
T2SWS, T2S & 6 & $\times$ & \checkmark & 6 & $\times$ & 42.69 & 54.10 \\
T2SWS, T2S & 6 & $\times$ & \checkmark & 0 & NA & 47.19 & 59.57 \\
\midrule
T2SWS, T2S & 0 & NA & \checkmark & 6 & \checkmark & 33.70 & 46.01 \\
T2SWS, T2S & 0 & NA & \checkmark & 6 & $\times$ & 46.07 & 56.50 \\
T2SWS, T2S & 0 & NA & \checkmark & 0 & NA & 46.08 & 59.03 \\
\midrule
T2S & 6 & $\times$ & \checkmark & 6 & \checkmark & 41.57 & 56.67 \\
T2S & 6 & $\times$ & \checkmark & 6 & $\times$ & 42.69 & 55.40 \\
T2S & 6 & $\times$ & \checkmark & 0 & NA & 52.80 & 69.63 \\
\midrule
T2S & 0 & NA & \checkmark & 6 & + & 39.33 & 55.58 \\
T2S & 0 & NA & \checkmark & 6 & -- & 48.31 & 62.83 \\
T2S & 0 & NA & \checkmark & 0 & NA & \textbf{60.67} & \textbf{72.35} \\
\bottomrule
\end{tabular}
\end{adjustbox}
\parbox{\linewidth}{\footnotesize Qwen2.5-Coder-Instruct-3B serves as the base model. \textbf{FS} denotes few-shot examples. \textbf{FS-C} is the number of few-shot examples, and \textbf{FS-R} indicates whether reasoning traces of few-shots are included in the prompt.  \textbf{T2S} refers to the fine-tuning dataset without schema context, whereas \textbf{T2SWS} includes filtered schema context. All fine-tuned models use LoRA (Rank = 32, Alpha = 32) with a learning rate of $1.0 \times 10^{-4}$, two training epochs, and an effective batch size of 8.}
\end{table}

\section{Discussion and Limitations}
\label{sec:discussion}
One limitation of the proposed framework arises when foreign-key constraints are absent or not explicitly defined in the database schema. Since the initial stage of SING-SQL relies on foreign-key relationships to guide schema partitioning, the lack of such constraints restricts the ability of the framework to construct accurate sub-schemas. In practice, this limitation can be mitigated by manually specifying foreign-key relationships, enabling the framework to proceed with schema partitioning even in databases where constraints are incompletely defined.

Another limitation of our framework arises when a set of columns within a table are semantically or functionally related and should be frequently queried together. Our current schema partitioning strategy applies a sliding window at the column level to generate sub-schemas. However, this approach does not consider inherent relationships among columns, which may result in separating fields that are typically used together. However, similar limitation can also be observed in the BIRD benchmark, despite being curated by domain experts. Appendix~\ref{appendix:seperation_columns_limitation} illustrates this issue with the Schools table in the California Schools database (part of the BIRD dev set), where administrator-related fields are split across different sub-schemas. Consequently, a question requiring all administrator email addresses may yield only partial results, reducing the semantic fidelity of question-SQL pairs.

A potential solution is to first identify and group semantically related columns before partitioning the columns via sliding window. This would help preserve meaningful column groupings within sub-schemas and avoid undesirable separation. However, such an approach generally requires domain knowledge and manual intervention, making it labor-intensive and not easily scalable.

Additionally, grouping semantically related columns would likely reduce the number of distinct sub-schemas and, in turn, the number of synthesized examples. To maintain data volume and diversity, this reduction could be addressed by generating larger amount of question-SQL pairs per sub-schema with variations in query focus or changing the parameters of the SING-SQL. Incorporating diverse natural language styles—similar to those used in OMNI-SQL~\cite{li2025omnisqlsynthesizinghighqualitytexttosql}—can further enhance the richness of the dataset without sacrificing structural coherence.

As future work, automated techniques could be developed to detect and preserve column groupings using LLMs, thereby mitigating this limitation without relying on manual effort or domain-specific expertise.

We do not study the effect of decoding hyperparameters such as temperature or top-p. These are fixed throughout our experiments to ensure consistency.

Reasoning traces are generated only through the divide-and-conquer prompting strategy. Exploring multiple reasoning paths could improve robustness and diversity but would substantially increase cost, and is therefore left as future work.

While schema linking could potentially be enhanced by incorporating LLM-based table filtering, optimizing linking performance is not the primary focus of this work. Instead, our goal is to demonstrate the effectiveness of different schema linking strategies when supported by synthetic data, rather than to develop the strongest possible schema linking system.

Additionally, we do not incorporate reinforcement learning (e.g., GRPO~\cite{DeepSeekMath}) as a post-training step. While such methods could further improve SQL generation by aligning outputs with execution results, we leave this as future work.


\section{Conclusion}
\label{sec:conclusion}
In this work, we introduced \textbf{SING-SQL}, a two-stage synthetic data generation framework tailored for in-domain Text-to-SQL tasks. By hierarchically partitioning target databases into diverse sub-schemas and generating SQL--Text pairs across multiple complexity levels, SING-SQL achieves both \textit{comprehensive schema coverage} and \textit{semantic alignment}. To ensure data quality, our framework integrates LLM-as-a-judge validation, executability checks, automatic repair, and column-focused balancing, resulting in datasets that are both high-coverage and structurally reliable.

We further released \textbf{SingSQL-LM}, a family of compact language models efficiently fine-tuned on the synthetic data produced by our framework. Extensive experiments on both the BIRD benchmark and our synthetic evaluation splits demonstrate consistent improvements in in-domain generalization. On the BIRD development subset, SingSQL-LM-3B-R64 attains 82.87\% Soft F1 and 73.03\% EX upper bound with 32 candidates, surpassing the best 3B-scale baseline by +16.21 points in Soft F1 and +12.36 points in EX. At the 1.5B scale, SingSQL-LM-1.5B-R64 improves over prior systems by +9.30 in Soft F1 and +4.49 in EX. On the synthetic evaluation splits, SingSQL-LM models further extend these gains. These results confirm high-coverage synthetic supervision not only benefits small-scale models but also establishes state-of-the-art performance among open models at comparable scales.

Notably, schema-free fine-tuning combined with schema-only inference yields the strongest results. This setting allows the model to internalize SQL patterns during training without being overly dependent on explicit schema supervision, while still leveraging schema context effectively at inference time. Our experiments show that synthetically generated few-shots and reasoning traces often destabilize performance once schema is provided, whereas schema-only inference consistently offers the most robust and scalable configuration for fine-tuned model. These findings underscore the practicality of schema-free training, as it reduces prompt complexity and cost at deployment while maintaining strong accuracy in domain-specific databases.

Beyond performance gains, our findings highlight the broader utility of synthetic database-specific data generation: it enables robust schema filtering with the help of LLM, enhances SQL generation, and offers a practical path for enterprises to adapt open-source LLMs without costly learning pipelines or large-scale manual annotation. Moreover, by combining parameter-efficient fine-tuning with compact models, we provide a viable solution for low-resource environments where annotated data or computational capacity is limited, making domain-specialized Text-to-SQL systems accessible even to organizations with constrained resources.

While our work addresses critical challenges in domain specialization, limitations remain. Current sub-schema partitioning may separate semantically related columns, and reasoning traces are generated through a single prompting strategy. Future work can explore automated column grouping, diverse reasoning styles, and reinforcement learning to further strengthen robustness.

Overall, SING-SQL establishes a \textit{scalable and database-agnostic paradigm} for generating high-quality in-domain data, providing a strong foundation for advancing Text-to-SQL translation in real-world enterprise settings.

\begin{acks}
  We thank TÜBİTAK ULAKBIM for granting access to the High Performance and Grid Computing Center (TRUBA), and Koç University High Performance Computing Center for providing GPU resources used in this work.
\end{acks}

\balance
\bibliographystyle{ACM-Reference-Format}
\bibliography{sample}

\appendix
\onecolumn 

\section{Flawed SQL-to-Text Translation}
\label{appendix:flawed_sql_to_text_translation}
When synthesizing Text-to-SQL pairs, one common strategy is to first generate SQL queries based on the database schema and subsequently produce corresponding natural language questions. This direction is often preferred as translating from SQL to natural language is considered a relatively simpler task compared to the inverse. However, SQL-to-text generation does not always yield high-quality question formulations. Accurate translation requires a deep semantic understanding of the database schema—specifically, what each column represents. In our experiments, we observed that even state-of-the-art large language models (LLMs) occasionally fail to capture these semantics accurately, resulting in flawed natural language questions. An illustrative example is provided in \autoref{fig:flawed_sql2text_example_1}, where a SQL query generated by Gemini-2.5-Pro selects the \texttt{enroll12} column from the \texttt{satscores} table. This column denotes the total number of students enrolled from 1st through 12th grade. However, the corresponding generated question incorrectly refers only to 12th-grade enrollment, demonstrating a semantic misinterpretation of the column’s meaning and resulting in low-quality data.

\begin{figure}[H]
\centering
\begin{tcolorbox}[
    colback=lightYellow,
    colframe=black,
    coltitle=black,
    sharp corners=south,
    rounded corners,
    boxrule=0.8pt,
    width=0.95\linewidth,
    arc=4pt,
    auto outer arc,
    fonttitle=\bfseries,
    enhanced,
    breakable,
    before skip=10pt,
    after skip=10pt,
    left=10pt,
    right=10pt,
    top=6pt,
    bottom=6pt
]
\raggedright
\framesectiontitle{SQL:} \\
\texttt{SELECT T1.AdmEmail1, T2.dname, T2.\redhighlight{enroll12} FROM schools AS T1 INNER JOIN satscores AS T2 ON T1.CDSCode = T2.cds INNER JOIN frpm AS T3 ON T1.CDSCode = T3.CDSCode WHERE T2.cname = 'Los Angeles' AND T3.'District Type' = 'Unified School District';}

\vspace{1em}

\framesectiontitle{Question:} \\
Show me the administrator's email, the district name, and the \redhighlight{12th-grade enrollment} for all schools located in Los Angeles county that are part of a Unified School District.
\end{tcolorbox}
\caption{Example of flawed SQL-to-Text translation where the model misinterprets column semantics.}
\label{fig:flawed_sql2text_example_1}
\end{figure}

\section{Illogical SQL-to-Text Pairs}
\label{appendix:illogical_sql_to_text_pairs}
In the process of generating Text-to-SQL training data, some SQL queries—though syntactically correct and semantically aligned with their corresponding questions—can still be illogical from an analytical or real-world perspective. These pairs are referred to as \textit{illogical SQL-to-text pairs}. Unlike flawed translations that arise from misinterpretations of column semantics, illogical pairs emerge when the intent or logic of the SQL query lacks analytical validity or practical utility.

One scenario involves performing numerical operations, such as computing an average, on values that are not inherently quantitative. This misstep often arises from a superficial interpretation of a column’s name or datatype (e.g., treating textual identifiers as numerically meaningful). While language models may generate SQL and natural language questions that are mutually consistent, the underlying query logic may not serve a meaningful purpose in practice.

An illustrative example is shown in \autoref{fig:illogical_sql2text_example_1}. Here, the SQL query attempts to calculate the \textit{average charter school identification number} for district types that have more than 500 schools. While technically executable and syntactically valid, the query is conceptually flawed—charter school identification numbers are unique IDs and not numerical attributes that should be averaged. Thus, the question generated from this query, despite being a literal translation, is illogical from an analytical standpoint.

\begin{figure}[H]
\centering
\begin{tcolorbox}[
    colback=lightYellow,
    colframe=black,
    coltitle=black,
    sharp corners=south,
    rounded corners,
    boxrule=0.8pt,
    width=0.95\linewidth,
    arc=4pt,
    auto outer arc,
    fonttitle=\bfseries,
    enhanced,
    breakable,
    before skip=10pt,
    after skip=10pt,
    left=10pt,
    right=10pt,
    top=6pt,
    bottom=6pt
]
\raggedright
\framesectiontitle{SQL:} \\
\texttt{SELECT "District Type", AVG(CASE WHEN "Charter School Number" IS NOT NULL THEN CAST("Charter School Number" AS REAL) ELSE NULL END) AS AverageCharterSchoolNumber FROM frpm GROUP BY "District Type" HAVING COUNT(CDSCode) > 500;}

\vspace{1em}

\framesectiontitle{Question:} \\
Which school district types have more than 500 schools, and what is their average charter school identification number (if available)?
\end{tcolorbox}
\caption{Example of an illogical SQL-to-Text pair. The SQL and question are semantically aligned but reflect an implausible analytical intent—averaging unique identifiers.}
\label{fig:illogical_sql2text_example_1}
\end{figure}

\section{Separation of Columns Highly Used Together Limitation Example}
\label{appendix:seperation_columns_limitation}
As discussed in Section~\ref{sec:discussion}, one drawback of our column-wise schema partitioning arises when semantically or functionally related columns are separated across different sub-schemas. In the California Schools dataset (from the BIRD dev set), the Schools table contains the following administrator-related columns:
\texttt{AdmFName1}, \texttt{AdmLName1}, \texttt{AdmEmail1}, \texttt{AdmFName2}, \texttt{AdmLName2}, \texttt{AdmEmail2}, \texttt{AdmFName3}, \texttt{AdmLName3}, \texttt{AdmEmail3}.

If not clearly stated, these columns should be often used together in queries that involve administrative contact information. However, because our current framework partitions columns using a sliding window without considering such semantic groupings, they may be split across multiple sub-schemas. However, similar limitation can also be observed in the BIRD benchmark, despite being curated by domain experts. 

Figure~\ref{fig:seperation_columns_problem_in_bird} illustrates an example of inconsistent handling of semantically related columns in the BIRD benchmark. The question asks for the common first names among school administrators, yet the provided SQL query only considers the first administrator (\texttt{AdmFName1}). This mismatch introduces ambiguity: either the SQL should aggregate across all administrator first-name columns, or the question should explicitly restrict the scope to the first administrator. A third possibility is to adopt a schema-level convention (e.g., defaulting to the first administrator unless otherwise specified), but such rules should be explicitly stated to avoid misinterpretation.
Figure~\ref{fig:seperation_columns_problem_in_synthetic_data} presents a synthetic example illustrating this limitation. The example question asks for administrator email addresses and unique identifiers for schools classified as 'County Community School'. Yet the corresponding SQL query only returns one email field (\texttt{AdmEmail3}), as the sub-schema does not include the other administrator email columns. This mismatch reduces the fidelity of the synthesized question-SQL pair and can negatively impact model training and evaluation when such examples appear in the train, dev, or test splits.

\begin{figure}[H]
\centering
\begin{tcolorbox}[
    colback=lightYellow,
    colframe=black,
    coltitle=black,
    sharp corners=south,
    rounded corners,
    boxrule=0.8pt,
    width=0.95\linewidth,
    arc=4pt,
    auto outer arc,
    fonttitle=\bfseries,
    enhanced,
    breakable,
    before skip=10pt,
    after skip=10pt,
    left=10pt,
    right=10pt,
    top=6pt,
    bottom=6pt
]
\raggedright

\framesectiontitle{Question:} \\
What are the two most common first names among the school administrators? Indicate the district to which they administer.

\vspace{1em}

\framesectiontitle{SQL:} \\
\texttt{SELECT DISTINCT T1.AdmFName1, T1.District FROM schools AS T1 INNER JOIN ( SELECT admfname1 FROM schools GROUP BY admfname1 ORDER BY COUNT(admfname1) DESC LIMIT 2 ) AS T2 ON T1.AdmFName1 = T2.admfname1;}

\end{tcolorbox}
\caption{Data example in Bird development set for the separation of semantically related columns problem.}
\label{fig:seperation_columns_problem_in_bird}
\end{figure}

\begin{figure}[H]
\centering
\begin{tcolorbox}[
    colback=lightYellow,
    colframe=black,
    coltitle=black,
    sharp corners=south,
    rounded corners,
    boxrule=0.8pt,
    width=0.95\linewidth,
    arc=4pt,
    auto outer arc,
    fonttitle=\bfseries,
    enhanced,
    breakable,
    before skip=10pt,
    after skip=10pt,
    left=10pt,
    right=10pt,
    top=6pt,
    bottom=6pt
]
\raggedright

\framesectiontitle{Sub-Schema:} \\
"frpm": [
    "CDSCode",
    "Percent (\%) Eligible FRPM (Ages 5-17)",
    "High Grade"
],\\
"satscores": [
    "cds",
    "AvgScrWrite",
    "NumTstTakr"
],\\
"schools": [
    "CDSCode",
    "EdOpsCode",
    "AdmEmail3"
]

\vspace{1em}

\framesectiontitle{SQL:} \\
\texttt{SELECT T1.AdmEmail3, T1.CDSCode FROM schools AS T1 WHERE T1.EdOpsCode = 'COMM';}

\vspace{1em}

\framesectiontitle{Question:} \\
What are the administrator email addresses and unique identifiers for schools classified as 'County Community School'?
\end{tcolorbox}
\caption{Synthetic Data Example for a separation of semantically related columns.}
\label{fig:seperation_columns_problem_in_synthetic_data}
\end{figure}

\newpage
\section{Schema Filtering Performance on Synthetic Data}
\label{appendix:schema_filtering_performance_on_synthetic_data}
\begin{table}[htbp]
\centering
\caption{Schema Filtering Performance on the Synthetically generated Text-to-SQL data for the California Schools database in Bird Dev Split. 
All methods retrieve few-shots using \emph{user-question keyword pairs}. 
``BM25'' retrieves examples using BM25; ``All'' uses all retrieved few-shots to construct filtered schema while ``Top6'' uses the top-6 most similar examples to the user question. ``+LLM'' applies LLM-based table filtering leveraging retrieval. ''TR'' and ''TP''represents Table Recall and Precision respectively. Similarly, ''CR'' and ''CP'' represents Column Recall and Precision respectively. ''SRR'' represents strict schema recall rate}
\label{tab:schema_filtering_performance_on_synthetic_data}
\begin{tabular}{lccccc}
\toprule
\textbf{Method} & \textbf{TR} & \textbf{TP} & \textbf{CR} & \textbf{CP}  & \textbf{SRR} \\
\midrule
\multicolumn{6}{c}{\textbf{Synthetic Train Dataset}}\\
\midrule
BM25-Top6              & 96.50 & 68.48 & 89.41 & 39.15 & 75.65 \\
BM25-All               & 97.14 & 64.96 & 93.10 & 16.88 & 87.78 \\
BM25-Top6 + LLM   & 97.09 & 65.92 & 92.08 & 45.59 & 83.66 \\
\midrule
\multicolumn{6}{c}{\textbf{Synthetic Dev Dataset}}\\
\midrule
BM25-Top6              & 95.51 & 65.71 & 86.94 & 37.01 & 71.97 \\
BM25-All               & 96.20 & 62.26 & 90.99 & 15.70  & 84.51 \\
BM25-Top6 + LLM   & 96.20 & 63.35 & 90.20 & 45.59 & 81.93 \\
\midrule
\multicolumn{6}{c}{\textbf{Synthetic Test Dataset}} \\
\midrule
BM25-Top6              & 96.12 & 66.63 & 89.03 & 37.78 & 75.71 \\
BM25-All               & 96.67 & 62.50 & 92.88 & 15.96 & 89.05 \\
BM25-Top6 + LLM   & 96.63 & 63.59 & 91.13 & 40.03 & 82.74 \\
\bottomrule
\end{tabular}
\end{table}

\section{Sub-Schema and Synthetic Data Statistics}
\label{appendix:ss_and_synth_data_statistics}

While synthetic Text-to-SQL data is generated only for the \textit{California Schools} database in the development set of the BIRD benchmark, Table~\ref{tab:ss_and_synth_data_statistics} reports statistics across multiple databases to illustrate how sub-schema generation varies with different parameter choices. For the experiments in this work, we specifically use the configuration in the first row of Table~\ref{tab:ss_and_synth_data_statistics}: the \textit{California Schools} database with 3 tables and 89 columns (an average of 29.66 columns per table). With a sliding window length of 3 and stride of 2, we obtain 2,249 sub-schemas and ultimately generate 39,734 synthetic examples.

\begin{table}[htbp]
\centering
\small
\caption{Sub-Schema and Synthetic Data Statistics}
\label{tab:ss_and_synth_data_statistics}
\resizebox{\textwidth}{!}{%
\begin{tabular}{lccccccccc}
\toprule
& \multicolumn{3}{c}{Database Properties} 
& \multicolumn{4}{c}{Sub-Schema and Synthetic Data Generation Parameters} 
& \multicolumn{2}{c}{Resulting Statistics} \\
\cmidrule(lr){2-4} \cmidrule(lr){5-8} \cmidrule(lr){9-10}
Database & 
\makecell{Table \\ Count} & 
\makecell{Column \\ Count} & 
\makecell{Avg. Cols \\ per Table} & 
\makecell{Table-Level Sub-Schema \\ Table Counts} & 
\makecell{Sliding \\ Window \\ Length} & 
Stride & 
\makecell{Min Col \\ Example \\ Count} & 
\makecell{Sub-Schema \\ Count} & 
\makecell{Synthetic \\ Data \\ Count} \\
\midrule
California Schools & 3 & 89 & 29.66 & [3, 2, 1] & 3 & 2 & 400 & 2249 & 39734 \\
California Schools & 3 & 89 & 29.66 & [3, 2, 1] & 3 & 1 & -- & 11420 & -- \\
Card Games & 6 & 115 & 19.16 & [3, 2, 1] & 3 & 2 & -- & 2938 & -- \\
Card Games & 6 & 115 & 19.16 & [3, 2, 1] & 2 & 1 & -- & 13352 & -- \\
Codebase Community & 8 & 71 & 8.875 & [4, 3, 2, 1] & 3 & 2 & -- & 3533 & -- \\
Codebase Community & 8 & 71 & 8.875 & [3, 2, 1] & 3 & 2 & -- & 1134 & -- \\
\bottomrule
\end{tabular}%
}
\end{table}

\newpage
\section{Column Coverage Comparison for California Schools Database (Bird vs. Synthetic Dataset)}
\label{appendix:column_coverage_comparison_california_schools}

\begin{figure*}[htbp]
  \centering
  \includegraphics[width=0.9\linewidth]{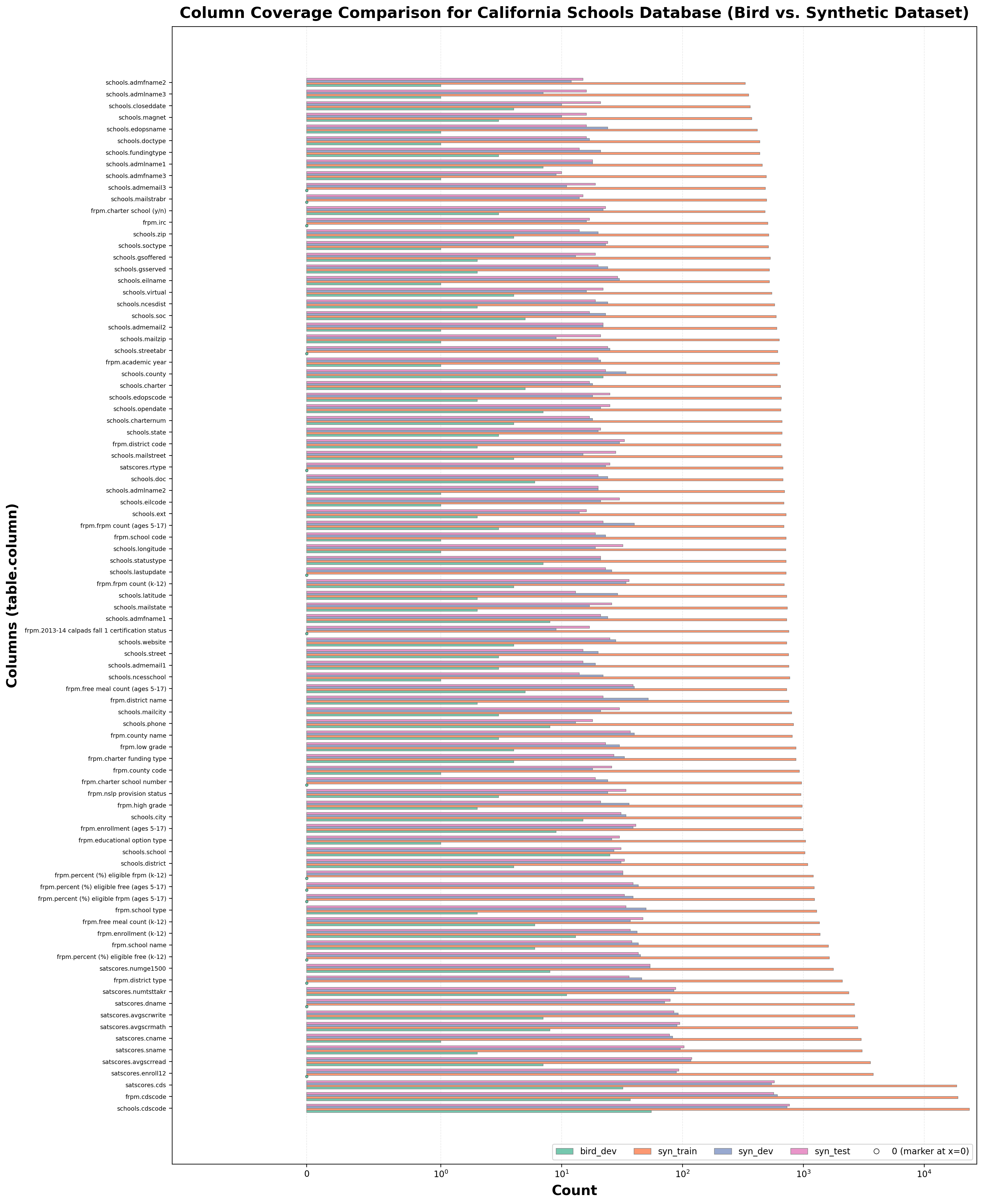}
  \caption{Column Coverage Comparison for California Schools Database (Bird vs. Synthetic Dataset)}
  \label{fig:column_coverage_comparison_california_schools}
\end{figure*}

\newpage
\section{Context Management Study}
\label{appendix:context_management_study}

\begin{table}[htbp]
\centering
\caption{Comparison of model performance across different training and inference contexts (candidate SQL count = 8). }
\label{tab:performance_comparison_of_models_with_various_contexts}
\renewcommand{\arraystretch}{1.00} 
\begin{adjustbox}{max width=\columnwidth, scale=1} 
\begin{tabular}{ccccccccccc}
\toprule
\multicolumn{3}{c}{\textbf{Training Context}} &  
\multicolumn{3}{c}{\textbf{Inference Context}} &  
\multicolumn{2}{c}{\textbf{Performance-R32}} &
\multicolumn{2}{c}{\textbf{Performance-R64}} \\
\cmidrule(lr){1-3} \cmidrule(lr){4-6} \cmidrule(lr){7-8} \cmidrule(lr){9-10}  
\textbf{Dataset} & \textbf{FS-C} & \textbf{FS-R} & \textbf{Schema} &
\textbf{FS-C} & \textbf{FS-R} & 
\textbf{EX UB} & \textbf{F1 UB} &
\textbf{EX UB} & \textbf{F1 UB} \\
\midrule
T2SWS & 6 & $\times$ & \checkmark & 6 & \checkmark & 40.44 & 52.55 & 38.20 & 45.65 \\
T2SWS & 6 & $\times$ & \checkmark & 6 & $\times$ & 33.70 & 46.38 & 44.94 & 57.93 \\
T2SWS & 6 & $\times$ & \checkmark & 0 & NA & 46.07 & 56.00 & 47.19 & 60.23 \\
\midrule
T2SWS & 0 & NA & \checkmark & 6 & \checkmark & 35.96 & 49.87 & 39.32 & 49.18 \\
T2SWS & 0 & NA & \checkmark & 6 & $\times$ & 40.45 & 53.29 & 44.94 & 55.07 \\
T2SWS & 0 & NA & \checkmark & 0 & NA & 51.69 & 63.20 & 51.68 & 62.81 \\
\midrule
T2SWS, T2S & 6 & $\times$ & \checkmark & 6 & \checkmark & 30.34 & 43.49 & 42.01 & 51.11 \\
T2SWS, T2S & 6 & $\times$ & \checkmark & 6 & $\times$ & 42.69 & 54.10 & 42.76 & 53.05 \\
T2SWS, T2S & 6 & $\times$ & \checkmark & 0 & NA & 47.19 & 59.57 & 43.82 & 55.63 \\
\midrule
T2SWS, T2S & 0 & NA & \checkmark & 6 & \checkmark & 33.70 & 46.01 & 43.82 & 52.33 \\
T2SWS, T2S & 0 & NA & \checkmark & 6 & $\times$ & 46.07 & 56.50 & 35.95 & 50.73 \\
T2SWS, T2S & 0 & NA & \checkmark & 0 & NA & 46.08 & 59.03 & 48.31 & 61.64 \\
\midrule
T2S & 6 & $\times$ & \checkmark & 6 & \checkmark & 41.57 & 56.67 & 46.06 & 64.57 \\
T2S & 6 & $\times$ & \checkmark & 6 & $\times$ & 42.69 & 55.40 & 49.44 & 61.40 \\
T2S & 6 & $\times$ & \checkmark & 0 & NA & 52.80 & 69.63 & 52.81 & 64.06 \\
\midrule
T2S & 0 & NA & \checkmark & 6 & + & 39.33 & 55.58 & 46.07 & 59.49 \\
T2S & 0 & NA & \checkmark & 6 & -- & 48.31 & 62.83 & 44.94 & 61.07 \\
T2S & 0 & NA & \checkmark & 0 & NA & \textbf{60.67} & \textbf{72.35} & \textbf{64.04} & \textbf{73.61} \\
\bottomrule
\end{tabular}
\end{adjustbox}
\parbox{0.75\linewidth}{\footnotesize Qwen2.5-Coder-Instruct-3B serves as the base model. \textbf{FS} denotes few-shot examples. \textbf{FS-C} is the number of few-shot examples, and \textbf{FS-R} indicates whether reasoning traces of few-shots are included in the prompt.  \textbf{T2S} refers to the fine-tuning dataset without schema context, whereas \textbf{T2SWS} includes filtered schema context.}
\end{table}

\newpage
\section{Schema Table Column Filtration Prompt Template}
\label{appendix:schema_table_column_filtration_prompt_template}

\begin{tcolorbox}[mybluebox, width=\textwidth]
\lstset{
    breaklines=true,
    basicstyle=\ttfamily\scriptsize, 
    columns=fullflexible,
    frame=none,
    backgroundcolor=\color{blue!5!white},
    xleftmargin=0pt,
    xrightmargin=0pt,
    showspaces=false,
    showstringspaces=false,
    keepspaces=true, 
    breakindent=0pt, 
    aboveskip=0pt,
    belowskip=0pt,
}

\begin{lstlisting}
### You are an excellent data scientist. You can capture the link between user question and database elements (columns). You determine the relevant database columns perfectly. Your objective is to analyze and understand the essence of the given question, single database table schema, examples and then select the relevant columns of a given table. 

### Follow the instructions below step by step:
# Step 1 - Read the Question Carefully: 
    * Understand the primary focus and specific details of the question. Identify named entities ( such as organizations, locations, etc.), technical terms, and other key phrases that encapsulate important aspects of the inquiry to establish a clear link between the question and the table columns. 
    * If a hint is given with the quesiton, review it. The hint provides specific information and directs attention toward certain elements relevant to the question and its answer. Use the hint to understand the relation between the question, the hint, and the given table columns. Always follow such logic explicitly.
# Step 2 - Analyze the Table Schema: 
    * You are given a schema of a single table in a database. Examine the table schema and detailed information about the columns to identify relevant columns that are pertinent to the question. 
    * Understand the meaning and purpose of the column, not just its name.
# Step 3 - Examine the Examples: 
    * Review each Text-to-SQL example that use the database elements the table come from. 
    * Analyze each question-SQL pair to understand the table columns and learn how they are used, in which contexts they are used.
    * Use examples to guide your decision, but **do not restrict yourself** to only columns seen in examples.
# Step 4 - Select Useful Columns: 
    * Consider each column one by one in detail to determine it is useful and required to answer the questions. When iterating through each column, write detailed reasoning why a column is necessary and useful or not. While evaluating a column, you can take the advantage from the examples.
        - At each new line, start with column name that you consider. Go through the information about the column and state the properties of the column. Then start reasoning whether it is related to the questoin or not.
    * For each column in the given table schema, ask yourself: Is this column directly necessary or indirectly helpful to answer the question? 
        - If yes, include it. 
        - If the question or hint specifies this column, include it. 
        - If it's part of a formula or computation, include it. 
        - If you're unsure, err on the side of including it.
    * When unsure about a column, prefer to include a column unless it explicitly contradicts the question or hint. Use examples to inform your judgment, but do not overfit to them.
    * Always follow the formula or calculation logic explicitly provided in the question or hint. 
        * **IMPORTANT** If the question and hint describe a formula using specific column names, you must select those exact columns, even if alternative or redundant columns (e.g., percentage, rate, average) are present in the schema. **Do not** assume that similar-looking columns satisfy the requirement.
    * If a question requires aggregate function on a column, you **must** select that column. 
        * Determine a question requires aggregate function, then think and elaborate on which column aggretation should be applied.
        * For example, if a user question needs counting (requiring COUNT aggregate function), then select a column on which aggregate function should be applied. You **must** list that column in your answer.
        * When the question requires unique values (e.g., IDs, URLs), the corresponding SQL query will use `SELECT DISTINCT`. Refer to column statistics ("Value Statics") to determine if `DISTINCT` is necessary.
    * If all columns of the table are irrelevant to the question, then return empty Python List for the selected_columns key in the response.
# Step 5: Output Format: 
    * Give your response in JSON format with two following keys: "reasoning" and "selected_columns" where value of a selected_columns **must** be a Python List and the value of the "reasoning" should be string reasoning on each column and their usefulness.

- **IMPORTANT NOTE:** Some columns might not be used in the examples given, but it can be necessary or useful. Although a column is not used in examples, it might be necessary or useful to answer the questions. Pay attention on the those columns that are not seen in the examples but important to answer the user question.
- **IMPORTANT NOTE:** You should output the column names as it is given in the Table Schema. 
- **IMPORTANT NOTE:** **If a hint given, you **MUST** add all columns mentioned or exist in the hint directly. This is a strict rule.** 
- **IMPORTANT NOTE:** **IN your reasoning part, start new line and evaluate a single column in detail by first writing its name, properties, your understanding about the column and then the reasons for why the column is relavent or irrelevant to the question by strictly following the instruction steps above one by one.** 

{TABLE_SCHEMA}

{EXAMPLES}

User Question:
{QUESTION_AND_HINT}

### Now, it is your turn.
### Respond in the JSON format as follows:
{{
    "reasoning": "Iterate over each column in the table and provide reasoning whether the column is useful and necessary to answer the user question. If you are not sure about the usefulness of a column, you should add it as well. While evaluation column, take advantage from the examples.",
    "selected_columns": ["column_1", "column_2", "column_3", ...]
}}
\end{lstlisting}
\end{tcolorbox}

\newpage
\section{SQL Generation Prompt Template}
\label{appendix:sql_generation_prompt_template}

\begin{tcolorbox}[mybluebox, width=\textwidth]
\lstset{
    breaklines=true,
    basicstyle=\ttfamily\scriptsize, 
    columns=fullflexible,
    frame=none,
    backgroundcolor=\color{blue!5!white},
    xleftmargin=0pt,
    xrightmargin=0pt,
    showspaces=false,
    showstringspaces=false,
    keepspaces=true, 
    breakindent=0pt, 
    aboveskip=0pt,
    belowskip=0pt,
}

\begin{lstlisting}
*** You are an expert SQL generator. Your task is to generate a SQL query for the given user question, considering *only* the {DB_ID} database.

### INTRUCTIONS:
* Understand the question:
    - Carefully read and interpret the user's natural language question.
    - Consider only the {DB_ID} database when analyzing the question.
    - Analyze the relation between question and database items.
* Determine the required database items:
    - In order to construct the SQL query to answer the user question, Determine which tables, columns, and values from the database are needed to answer the question.
    - Analyze the relations between selected tables and columns 
* Apply Logical Filtering:
    - Identify the required filtering conditions, aggregations, groupings, window functions, orderings or limit needed to fulfill the intent of the question.
* Construct the SQL query:
    - Construct a valid and executable SQLite SQL query that directly answers the question using only the relevant parts of the schema.


{AUGMENTATION}

### Generate SQLite SQL query for the following question considering **only** {DB_ID} database.
*** Question ***
{QUESTION}

### Respond in the following format:
<reasoning>
Analysis about the question purpose and relation between database items. Steps to answer the user question and create correct SQL query in detail. Very detailed reasoning and logic to create correct SQLite SQL query. The reasons for selecting database items (tables and columns). Filters, aggregations and window functions that should be utilized and applied with their reasoning.
</reasoning>
<answer>
Generated SQLite SQL query to answer the question.
</answer>

### Now is your turn to respond in the above format:
\end{lstlisting}
\end{tcolorbox}

\end{document}